# FakeSpotter: A content and strategy agnostic Viral Misinformation Detection Tool

Giovanni Spitale[1,] 0000-0002-6812-0979

Federico Germani[1,2*] 0000-0002-5604-0437

1: ITE Lab, Institute of Biomedical Ethics and History of Medicine, University of Zurich, Zurich, Switzerland.

2: Institute for Data Science and Artificial Intelligence, Boğaziçi University, Istanbul, Türkiye

*: Corresponding Author. federico.germani@ibme.uzh.ch

## Abstract

Misinformation detection tools often rely on binary true/false classifications or models trained on historical examples, limiting their usefulness when novel misleading narratives emerge. Here, we present FakeSpotter, a content- and strategy-agnostic tool designed to estimate the viral misinformation risk of textual content by measuring structural "fingerprints" of misinformation rather than directly adjudicating truthfulness. FakeSpotter operationalizes a theory-driven framework across linguistic, narrative, logical, and critical-thinking dimensions, using repeated LLM assessments and domain-specific logistic regression classifiers for short and long texts. In a labelled corpus of 764 texts from social media and FakeNewsNet, FakeSpotter achieved macro F1 scores of 0.788 for short texts and 0.793 for long texts on a held-out test set. FakeSpotter's interpretive layer provides explainable outputs through feature-based scores, signal agreement, and a caution index—and can be used for social listening. These findings suggest that identifying the structural fingerprints of misinformation can support early, explainable, and human-supervised assessment of potentially viral misinformation.

## Background

The digital information environment has changed faster than our tools for understanding it. Disinformation—false or misleading information created, presented, and disseminated with the intention to cause harm [1]—can now be generated at scale, refined in seconds, localized across languages, and tailored to specific audiences through the expanding capabilities of large language models (LLMs) [2]. Arguably, these systems are not inherently malign, but they have introduced a new asymmetry: the cost of producing persuasive falsehoods has collapsed, while the cost of verifying and contextualizing information has remained high. All of this is unfolding within already fragile information ecosystems [3]. By contrast, misinformation refers to false or misleading information shared without harmful intent [1]. In this paper, when intentionality is not central to the analysis, we use the term misinformation for simplicity to refer broadly to misleading, false information.

Current approaches to misinformation detection can be broadly categorized into several distinct methodologies, each with specific strengths and significant limitations. Most available tools, however, fall prey to two overarching problems. Here we argue that the first issue is the binary characterization of information as real versus fake, or true versus false, which overlooks the complexity of misleading claims with viral potential. The second issue, we contend, is closely linked to the first, and is the heavy reliance on machine learning (ML) and deep learning (DL) systems trained on past data. Arguably these approaches struggle when confronted with emerging forms of misinformation, especially around novel topics whose existence could not have been anticipated in advance [4]. Computational detection has most frequently been framed as a supervised classification task, where models are trained on labeled datasets to distinguish between legitimate and deceptive content. Early and widely used ML algorithms include Support Vector Machines (SVM), Random Forests, Naive Bayes, and Logistic Regression [5–8]. These methods typically rely on manual feature extraction, such as specific linguistic cues or stylistic markers [4]. More advanced DL systems employ Convolutional Neural Networks (CNNs), Recurrent Neural Networks (RNNs), and Transformer-based models such as BERT or DeBERTa, which automatically learn more intricate patterns from data [9–12]. Graph Neural Networks (GNNs) have also emerged as a technique designed to analyze propagation patterns and network structures through which disinformation spreads [4]. More recently, various LLMs have been used for zero-shot or few-shot detection, as well as task-specific fine-tuning [13–15].

Yet reliance on historical data creates a structural weakness. Emerging falsehoods often concern previously unforeseen events, crises (e.g., pandemic or wars), and are characterized by information voids, which constitute fertile grounds for the proliferation of misinformation [3]. In such contexts, the relevant patterns, features, and topics have not been digested by models and learned by algorithms. As a result, detection systems may fail precisely when they are needed most. The alternative remains manual fact-checking by domain experts, which is too slow and resource-intensive to keep pace with the speed and abundance of circulating information online [16].

Within this broader landscape, several methodological families can be distinguished. Linguistic and style-based methods rely on the hypothesis that misinformation possesses recognizable writing signatures, such as emotionally manipulative tones or distinctive lexical patterns. While effective in identifying sensationalism, they often overfit to specific datasets and can create a false sense of security when encountering domain shifts or new topics [4]. Context and propagation-based methods instead analyze how information spreads across networks, using features such as user interaction patterns or diffusion graphs [4,17,18]. These approaches are often more resilient to direct content manipulation, but they are poorly suited for early detection because they require time and the accumulation of social metadata. Knowledge-based and fact-checking approaches compare claims against external databases or existing debunks. While highly accurate for previously verified claims, they struggle with novel misinformation that has not yet been recorded or assessed by experts [4,19]. LLM-based supervised classifiers represent a further development [20], but they introduce additional concerns. Their lack of explainability limits the possibility of providing clear rationales for why a piece of content was flagged [4,21], and their performance depends on the availability in the training data of existing debunks and accurate information about the content under scrutiny.

All these issues are not only technical, but also ethical. Guidance in infodemic management and the responsible use of AI in public health has stressed that explainability is essential to understand outputs, minimize bias, and preserve trust, especially during emergencies, when accurate information may itself be scarce [22–24].

One recent attempt to address at least the problem of binary classification proposes treating misinformation not as a fixed label, but as a measurable risk. Ruani et al. developed the Misinformation Risk Assessment Model (MisRAM), which conceptualizes misinformation through gradations of potential harm, exposure, and severity [25]. Rather than asking whether a claim is simply true or false, this approach shift attention toward the degree of risk posed by misleading content and its capacity to generate societal harm.

Building on this line of reasoning, here we present FakeSpotter, a content and strategy agnostic Viral Misinformation Detection tool designed to address these limitations simultaneously. First, in line with the perspective advanced by Ruani et al., the tool does not treat misinformation as a binary category. Not every false statement carries the same relevance or societal danger. A person may falsely claim that it is raining in Paris while residents can immediately verify that the sky is clear. Such a falsehood has little realistic potential to spread widely, persuade audiences, or generate meaningful harm. The central problem, therefore, is not whether a statement is false, but whether it has high viral, manipulative potential. FakeSpotter shifts attention toward precisely these higher-risk forms of misleading content. Its purpose is not to identify every inaccurate statement, but to detect information that contains characteristics associated with successful misinformation, i.e., claims that can circulate rapidly and shape beliefs before corrections emerge. Here we show that FakeSpotter relies on the concept of fingerprints of misinformation [26], i.e., stable characteristics that tend to recur across misleading narratives regardless of topic, language, or immediate context. While the surface subject may change (from health crises to elections, wars, migration, or financial panic) the persuasive mechanics often remain similar. We argue that effective misleading content frequently draws on recurring elements, because it has to leverage human vulnerabilities to succeed. And regarding the explainability problem highlighted above [21], FakeSpotter addresses this challenge by asking an LLM-powered system to focus on identifying and measuring the fingerprints of misinformation associated with misleading content, instead of asking an LLM to determine whether a text contains misinformation or not. We contend that binary verdicts often obscure the rationale behind a classification and may reproduce biases embedded in available datasets. By contrast, measuring observable fingerprints enables more explainable and transparent outputs.

# Methods

## 1. Theoretical framework

FakeSpotter is grounded in the work of Redaelli et al., who proposed a content-agnostic theory of misinformation [26]. Drawing on Roman Jakobson's model of language functions [28], Vladimir Propp's morphology of the folktale [29], classical logic reasoning and logic fallacies, and critical thinking research [30], Redaelli et al.'s framework identifies four layers of structural features — linguistic, narrative, logical, and critical thinking— that characterize effective misinformation independently of its topic, medium, or source. The central claim is that manipulative misinformation is structurally constrained: regardless of what it is about, it exploits a recurring and measurable repertoire of rhetorical, narrative, and inferential patterns. This property makes it possible to design detection tools that are preventive rather than reactive— operating on form rather than on content.

FakeSpotter operationalizes this framework as a computational pipeline. Each input text is assessed along 19 theoretically grounded dimensions (Table 1), organized into four analytical layers: Jakobson-inspired linguistic functions (J), Propp-inspired narrative archetypes (P), logical fallacies (L), and critical thinking indicators (C). An additional direct misinformation signal (DIS) and an irony/sarcasm detector (LOL) complete the scoring space. These 19 raw scores, produced by a LLM, are fed into domain-specific logistic regression classifiers to yield a final risk index and an interpretive output.

*Table 1. The 19 scoring dimensions of FakeSpotter, grouped by analytical layer.*

| Layer | Code | Description |
|---|---|---|
| Linguistic (Jakobson) | J-EMO | Emotive function: emotional language and affective loading |
| | J-CON | Conative function: directive or manipulative intent |
| | J-REF-S | Referential support: references used to substantiate claims |
| | J-REF-C | Referential challenge: references used to contest or undermine claims |
| Narrative (Propp) | P-ANT | Antagonist framing: opposing force portrayed as threatening or secretive |
| | P-HEL | Helper association: author or in-group framed as benevolent |
| | P-TUR | Turning point: critical escalation or narrative shift |
| | P-PUN | Punishment: negative actors framed as receiving consequences |
| Logical fallacies | L-NAT | Appeal to nature |
|  | L-CAR | Bandwagon fallacy (argumentum ad populum) |
|  | L-BUR | Burden of proof fallacy |
|  | L-OCC | Failure to follow Occam's razor |
|  | L-INC | Inconsistency / internal contradictions |
|  | L-FAL | Lack of falsifiability |
|  | L-CON | Conjunction fallacy |
| Critical thinking | C-CCC | Causation–correlation conflation |
|  | C-STA | Insufficient statistical reasoning |
| Auxiliary signals | DIS | Direct signals (false, misleading, or deceptive information or intent) |
|  | LOL | Irony / sarcasm detection |

# 2. System architecture

FakeSpotter is organized into four functional layers: (i) an API layer, which dispatches each text to the LLM and collects raw scores; (ii) a scoring layer, which standardizes the raw scores and applies frozen logistic regression classifiers to produce a calibrated risk index; (iii) an interpretation layer, which translates numeric output into a structured, human-readable assessment; and (iv) a persona, which is the system prompt that instantiates the theoretical framework in the LLM.

## 2.1. LLM scoring

Each text is submitted to GPT-4o mini (gpt-4o-mini-2024-07-18; OpenAI) via the Responses API. The model receives a structured system prompt (the persona) that defines its role and output format, as well as the target text as the user message. Temperature is set to 0 to minimize within-run stochasticity. To estimate assessment reliability, each text is evaluated across multiple independent runs (N = 3 by default); the mean and standard deviation across runs are computed for each of the 19 dimensions. A run is accepted only if the returned JSON contains exactly the 19 expected keys with numeric values in [0, 1]; malformed outputs trigger automatic retry (up to 3 attempts with exponential backoff). The persona instructs the model to produce a single valid JSON object with no surrounding text. Scoring rubrics for each dimension are stated in natural language within the persona, directly encoding the operationalizations of the theoretical framework. Scores for all dimensions are real-valued in [0, 1] and rounded to two decimal places.

## 2.2. Logistic regression classifiers

Raw LLM scores are transformed into a binary misinformation risk estimate via two domain-specific logistic regression classifiers—one for short texts (≤ 80 words) and one for long texts (> 80 words)—trained on the labelled dataset described in Section 3. The 80-word threshold was chosen to separate social media posts and brief statements, which constitute most of the short-domain corpus, from longer news and opinion pieces. This distinction was necessary because shorter texts exhibit different characteristics and disinformation fingerprints than longer texts. While longer texts are generally more likely to contain at least some disinformation-related fingerprints, their greater length may also dilute the relative salience of those fingerprints.

Each classifier is implemented as a sklearn Pipeline comprising a StandardScaler followed by LogisticRegression with ElasticNet regularisation (l1_ratio = 0.5, solver = saga, max_iter = 3,000). The regularization hyperparameter C is selected via 5-fold stratified cross-validation optimizing AUC-ROC. Only features that survive Bonferroni correction (α = 0.05, Mann–Whitney U test) within each domain are passed to the corresponding classifier, reducing the risk of fitting noise and limiting multicollinearity. At runtime, the fitted scaler parameters and logistic regression coefficients are loaded from frozen JSON files (model_short.json, model_long.json), eliminating any dependency on sklearn at inference time and ensuring exact reproducibility across environments. The sigmoid output of the linear combination is scaled to [0, 100] to yield the risk index (lr_score), where 0 corresponds to information with no viral potential, and 100 to misleading information with a high virality risk.

## 2.3 Auxiliary signals and caution index

In parallel with the LR classifier, FakeSpotter computes two auxiliary signals. The direct misinformation signal (DIS score) is the LLM's unmediated holistic assessment of whether the evaluated text is false or misleading, and serves as an independent estimate of risk. The irony/sarcasm signal (LOL score) flags texts in which literal and intended meaning are systematically misaligned, a condition that can cause both the LLM assessor and the LR classifier to produce unreliable scores. The relationship between lr_score and dis_score is characterised via a four-quadrant signal agreement schema. When both signals agree that a text is authentic (both < 50) or misinformation (both ≥ 50), the output is classified as agreement_true or agreement_false respectively. When the signals diverge — the LR classifier flags manipulative linguistic form while the holistic DIS score does not (divergence_factcheck), or vice versa (divergence_content) — a targeted alert is raised, provided the absolute gap between the two signals exceeds 20 points. A caution index (caution_score, [0, 100]) aggregates three sources of interpretive uncertainty as an equal-weight mean: (i) weighted cross-run instability of LLM scores, where per-feature standard deviations are weighted by normalized absolute LR coefficients; (ii) the LOL score; and (iii) the absolute disagreement between lr_score and dis_score after a tolerance band of 0.1 (i.e. differences between 0 and 0.1 are considered as 0). The caution index does not modify the risk estimate; it signals to the user that the output should be interpreted with additional care.

## 2.4 Risk bands and interpretation layer

The continuous lr_score is discretized into six ordinal risk bands derived from calibration curves on the held-out test set (Table 2). Each band is assigned a qualitative verdict, ranging from “Very likely true information with no meaningful viral misinformation risk” to “Very likely misinformation with high viral spread potential”, that, together with the caution index and signal agreement quadrant, constitutes the user-facing output. The interpretation layer supports multilingual output.

*Table 2. Risk band cutoffs and qualitative verdicts.*

| Band | lr_score range | Verdict |
|---|---|---|
| 0 | < 15 | Very likely true information with no meaningful viral misinformation risk |
| 1 | 15 – 34 | Likely true information with low viral misinformation risk |
| 2 | 35 – 49 | Uncertain content, leaning true, with limited misinformation risk |
| 3 | 50 – 64 | Uncertain content, leaning misinformation, with moderate viral risk |
| 4 | 65 – 79 | Likely misinformation with elevated viral risk |
| 5 | ≥ 80 | Very likely misinformation with high viral spread potential |

# 3. Data

## 3.1. Dataset construction

A labelled corpus was assembled from two sources. The primary source is based on a hand-curated dataset we built consisting of 653 texts collected from social media platforms, predominantly from Twitter/X and LinkedIn, covering a range of topics including public health, climate change, and political discourse. Texts were labelled as false (misinformation) or true (accurate) by the authors; texts labelled as 'undetermined' (uncertain classification) were excluded, yielding a total of 280 false and 250 true items. The secondary source is FakeNewsNet [31], a publicly available benchmark dataset of fact-checked news articles from PolitiFact. The two sources were merged into a single corpus of 768 texts. All texts were classified into two length domains: short (≤ 80 words) and long (> 80 words).

## 3.2. API assessment and checkpoint procedure

Each text in the merged corpus was submitted to the API assessment pipeline described in 2.1. Each row of the checkpoint file corresponds to one text and stores the per-feature mean and standard deviation across the $N_{assessments}$ = 3 assessment runs, together with the text identifier, label, domain classification, word count, source, and an error field. Rows for which the API call failed (due to network errors, malformed JSON output that could not be reconciled after three retry attempts, or all-zero score vectors indicative of a silent processing failure) were flagged with a non-null error value. Prior to any downstream analysis, all flagged rows ($N_{errors}$ = 4) were removed from the dataset. Only texts with complete, valid score vectors across all 19 dimensions were retained for feature selection, model training, and evaluation ($N_{tot}$ = 764). The final sample sizes reported in Table 3 reflect the corpus after this cleaning step.

*Table 3. Dataset composition by domain and label.*

| Domain | False | True | Total |
|---|---|---|---|
| Short (≤ 80 words) | 259 | 212 | 471 |
| Long (> 80 words) | 136 | 157 | 293 |
| Total | 395 | 369 | 764 |

## 3.3. Train/test split

The corpus was partitioned into a training set (80%; n = 611) and a held-out test set (20%; n = 153) using stratified random sampling. The stratification key was the cross of domain and label (four strata: false_short, true_short, false_long, true_long), ensuring proportional representation in both partitions. No text from the test set was used at any stage of feature selection, model training, or hyperparameter optimization. The random seed was fixed to ensure reproducibility.

# 4. Model calibration and validation

## 4.1. Feature selection and model training

Within the training set, Mann–Whitney U tests were run separately for each domain and each of the 19 LLM-derived feature means, comparing distributions between false and true texts. P-values were Bonferroni-corrected for the 19 simultaneous comparisons within each domain ($\alpha_{corrected}$ = 0.05 / 19 ≈ 0.0026). Only features surviving this threshold were included in the corresponding classifier. Effect sizes were quantified using rank-biserial correlation and Cohen's d. Variance inflation factors (VIF) were inspected to detect multicollinearity among selected features; features with VIF > 10 were flagged for review. Logistic regression models with ElasticNet regularization were trained separately for each domain on the Bonferroni-selected features. Hyperparameter C was optimized via 5-fold stratified cross-validation (scoring: AUC-ROC). Feature weight stability was assessed by re-fitting the model on each of the five folds and computing the mean and standard deviation of coefficients across folds. Learning curves were generated to evaluate convergence and detect overfitting.

## 4.2. Evaluation on the held-out test set

Model performance was evaluated on the held-out test set independently for each domain. Primary metrics were AUC-ROC and AUC-PR (average precision), selected for their threshold-independence and robustness to class imbalance. Bootstrap 95% confidence intervals for AUC were computed using 1,000 stratified resamples. Secondary metrics at a 0.5 probability threshold included accuracy, macro F1, per-class F1, Brier score, and Cohen's $\kappa$. Confusion matrices were computed to characterize the error structure. SHAP values (LinearExplainer, exact Shapley values for logistic regression) were computed on the test set to quantify and visualize the contribution of each feature to individual predictions, and are presented in the Supplementary Material.

## 4.3. Validation of the interpretive system

The interpretive layer (risk bands, signal agreement quadrants, and caution index) was validated through four pre-specified tests conducted on the scored test set. Test 1 (Calibration per band) assessed whether each band's observed false rate fell within its expected range and was computed with 95% Wilson confidence intervals. Test 2 (Monotonicity) tested whether the false rate increased monotonically across bands 0–5 using Kendall's $\tau$. Test 3 (Uncertainty honesty) examined whether texts assigned to uncertain bands (2–3) showed higher misclassification rates than texts in confident bands (0–1, 4–5), tested via $\chi^2$ on the 2 × 2 contingency table of band type × classification outcome. Test 4 (Signal agreement informativeness) tested whether the signal_agreement quadrant was independently predictive of misclassification using a $\chi^2$ test of independence and per-quadrant accuracy estimates with Wilson confidence intervals.

# Results

## 1. Feature analysis on the training set

### 1.1. Score distributions and discriminative power

All 19 LLM-derived features were evaluated on the training set (80% of the corpus) separately for short and long text domains. Figure 1 displays score distributions for each feature, stratified by label. Bonferroni-corrected Mann–Whitney U tests revealed that in the short domain all 19 features showed statistically significant separation between false and true texts (all $p_{Bonferroni} < 0.001$). In the long domain, 14 of 19 features reached significance; L-NAT, L-CAR, L-BUR, LOL, and L-CON did not survive correction ($p_{Bonferroni} \geq 0.05$), consistent with their lower rank-biserial r values. Across both domains, the features with the largest effect sizes were those reflecting inconsistency and failures of critical thinking (L-INC, C-CCC), direct misinformation signals (DIS), and the absence of referential support (J-REF-S). In the long text domain, the narrative turning-point signal (P-TUR) emerged as a particularly strong discriminator ($r = 0.44$, $p < 0.001$), alongside J-EMO and P-PUN, suggesting that longer misinformation texts rely more heavily on narrative escalation and emotional-directive language.

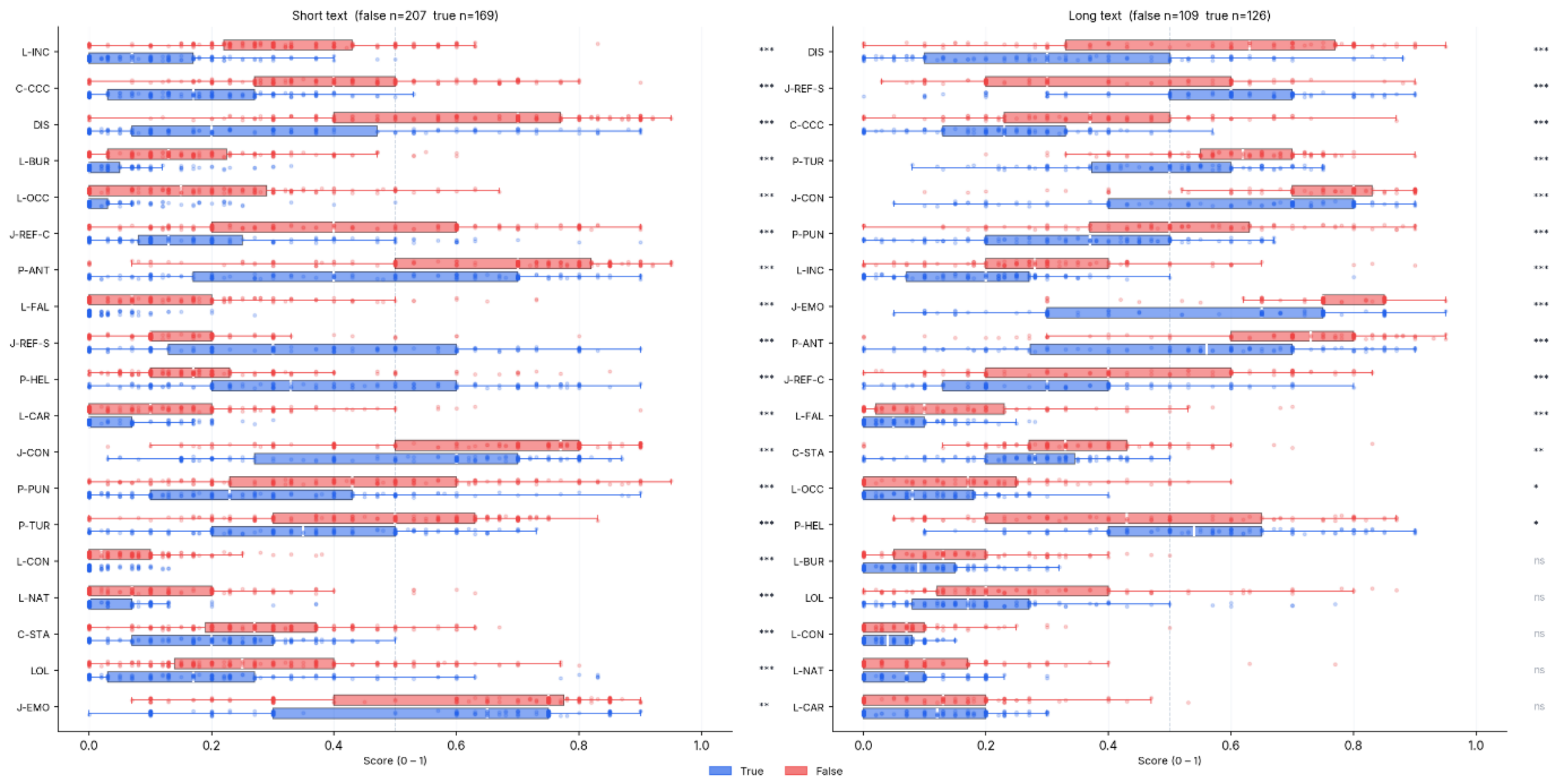

*Figure 1. Feature score distributions by label, training set (80%). Boxplots with strip plots for all 19 features, sorted by |rank-biserial r|. Stars indicate Bonferroni-corrected significance (Mann–Whitney U): *** p < 0.001, ** p < 0.01, * p < 0.05, ns = not significant. Left panel: short texts (≤ 80 words); right panel: long texts (> 80 words).*

## 1.2 Feature correlations and multicollinearity

Pearson correlation matrices revealed pervasive inter-feature correlations in both domains (Figures 2a–b). Particularly high correlations (r > 0.80) were observed between J-EMO and J-CON (short text: r = 0.92; long text: r = 0.85), between J-CON and J-REF-C (short text: r = 0.85; long text: r = 0.82), between L-BUR and L-OCC (short text: r = 0.87; long text: r = 0.81), and between DIS and C-CCC (short text: r = 0.84; long text: r = 0.87). These patterns reflect the structural co-occurrence of linguistic, logical, and critical thinking fingerprints in misinformation.

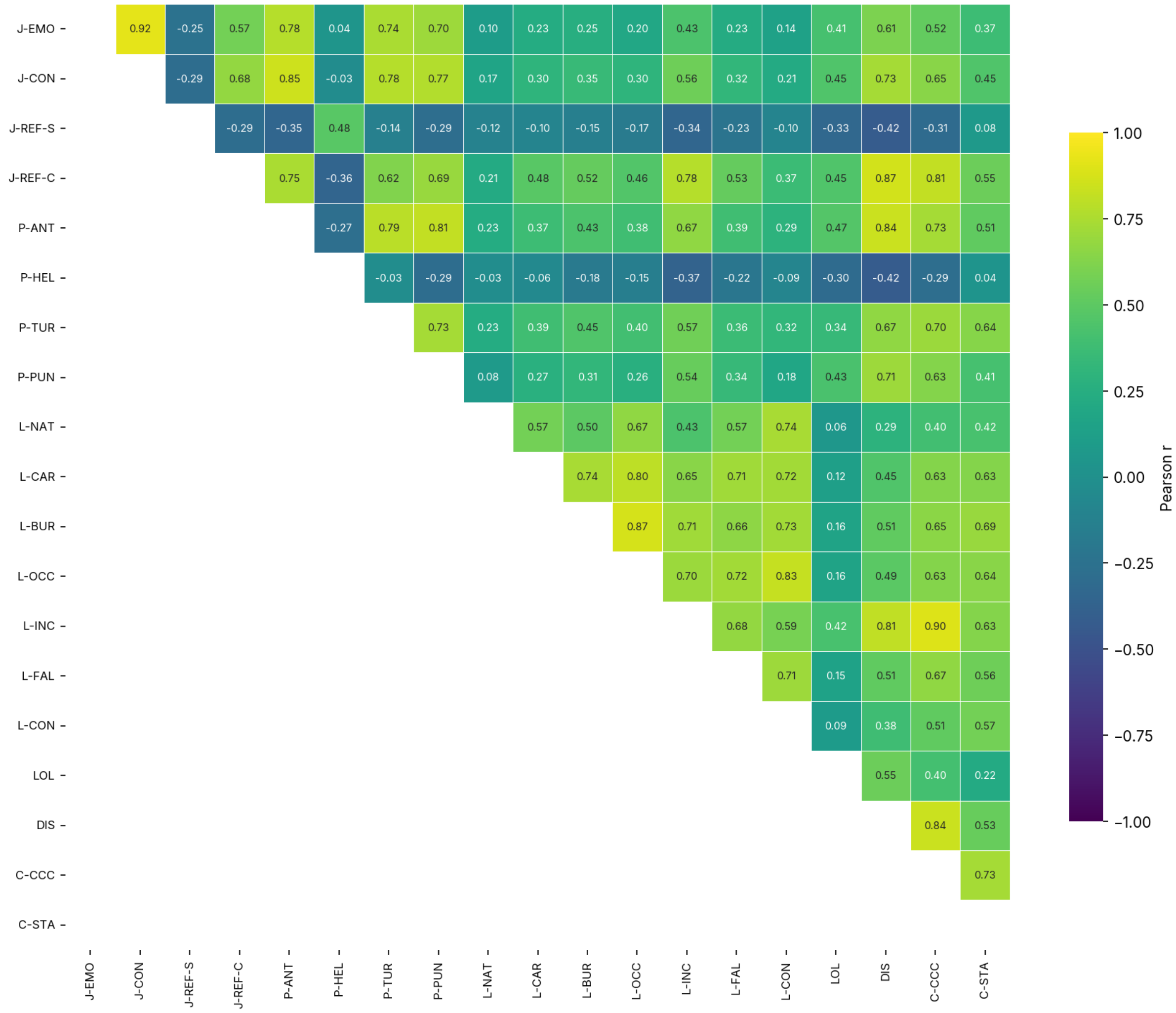

*Figure 2a. Pearson correlation matrix, short texts, training set. Upper triangle only; values represent Pearson's r.*

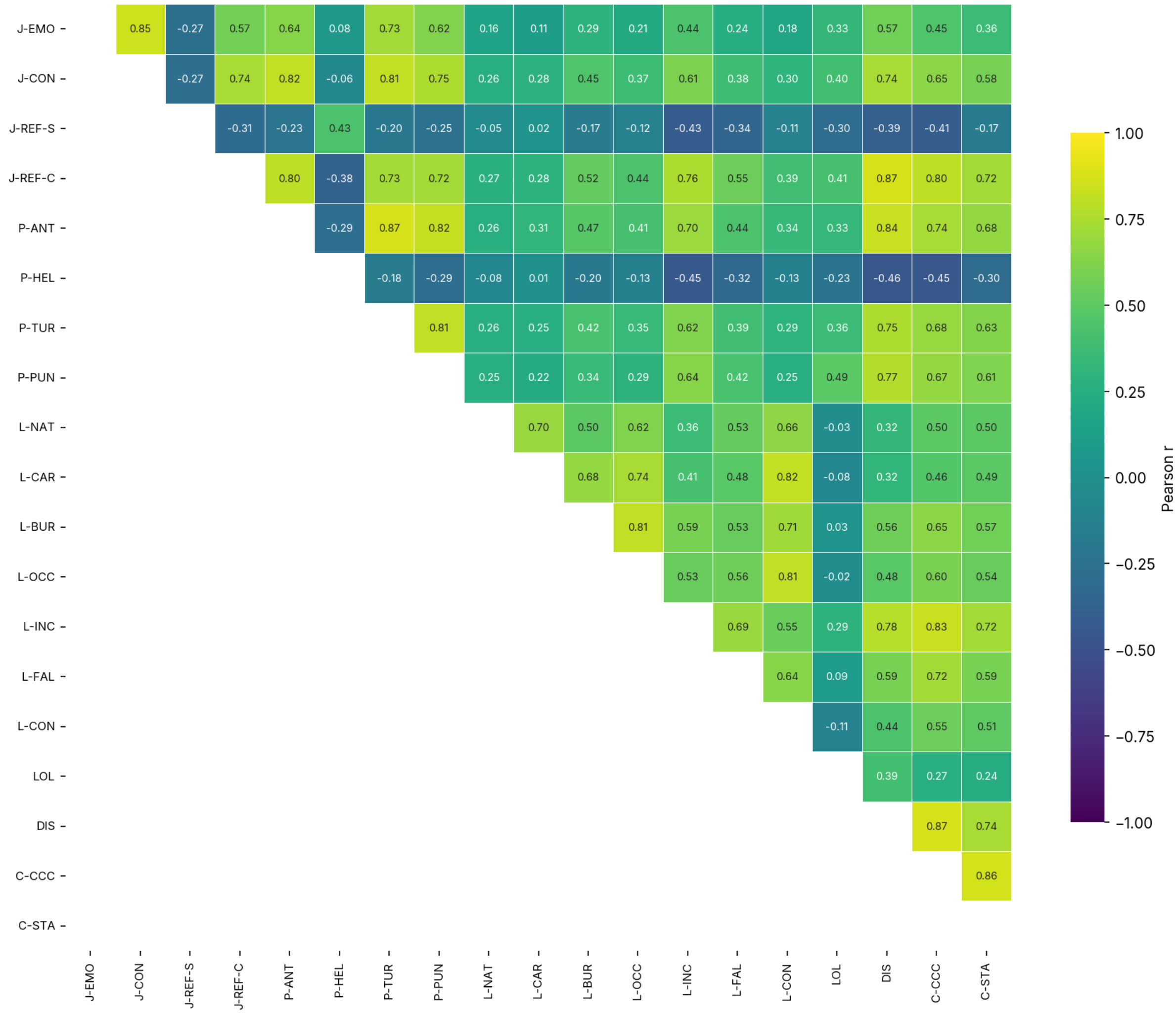

*Figure 2b. Pearson correlation matrix, long texts, training set..*

Variance inflation factors (VIF) confirmed the severe multicollinearity in the full 19-feature space, with J-CON reaching VIF = 78.15 (short text) and 79.71 (long text); Figures 3a–b). Three features (L-NAT, L-FAL, and LOL) remained below VIF = 5 in both domains, indicating negligible multicollinearity. Multicollinearity among selected features is addressed downstream by ElasticNet regularization, which applies L1/L2 shrinkage to correlated coefficients in the logistic regression stage.

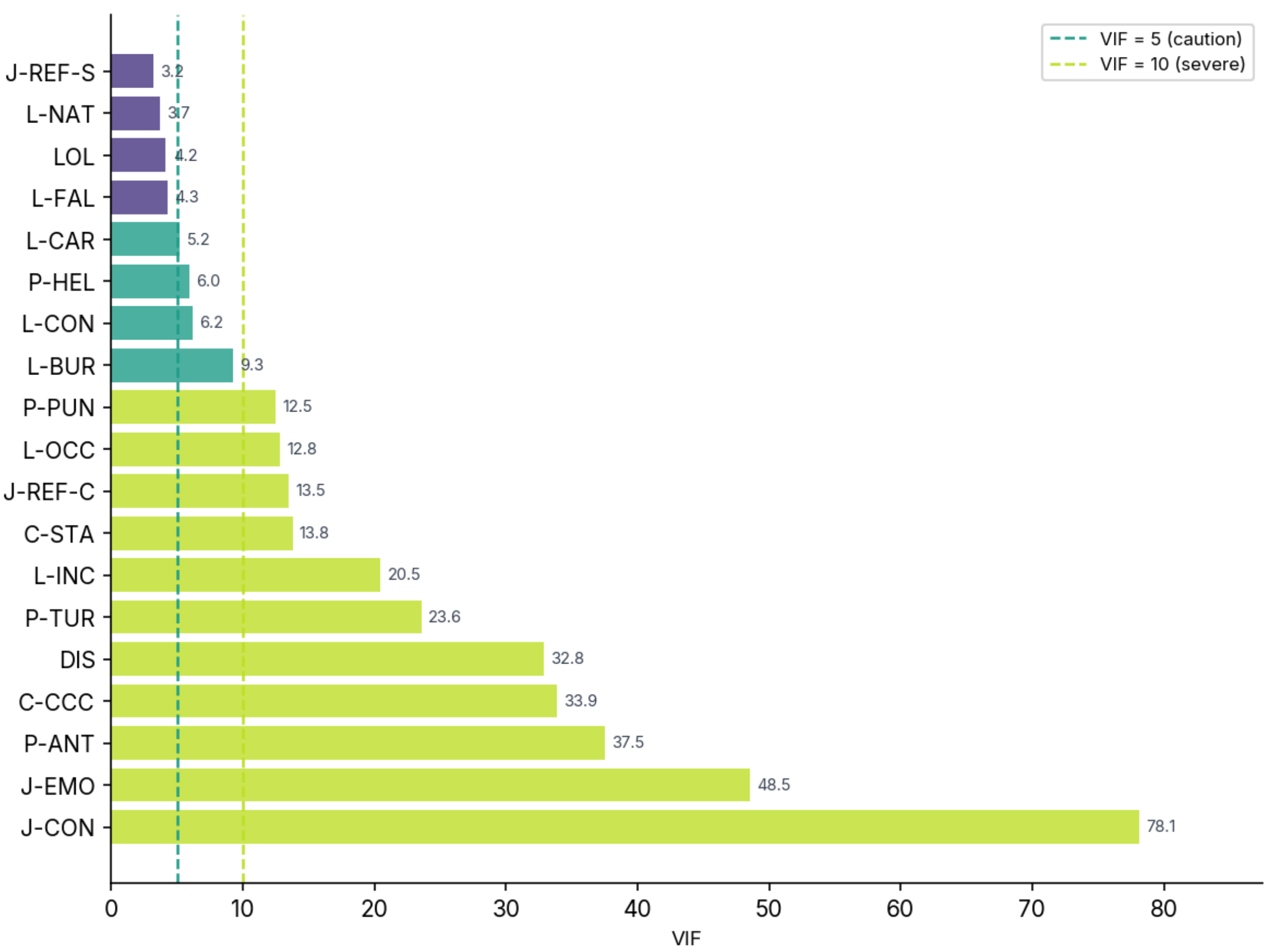

*Figure 3a. Variance inflation factors (VIF), short texts, full feature set. Orange dashed line: VIF = 5 (caution threshold); red dashed line: VIF = 10 (severe multicollinearity threshold).*

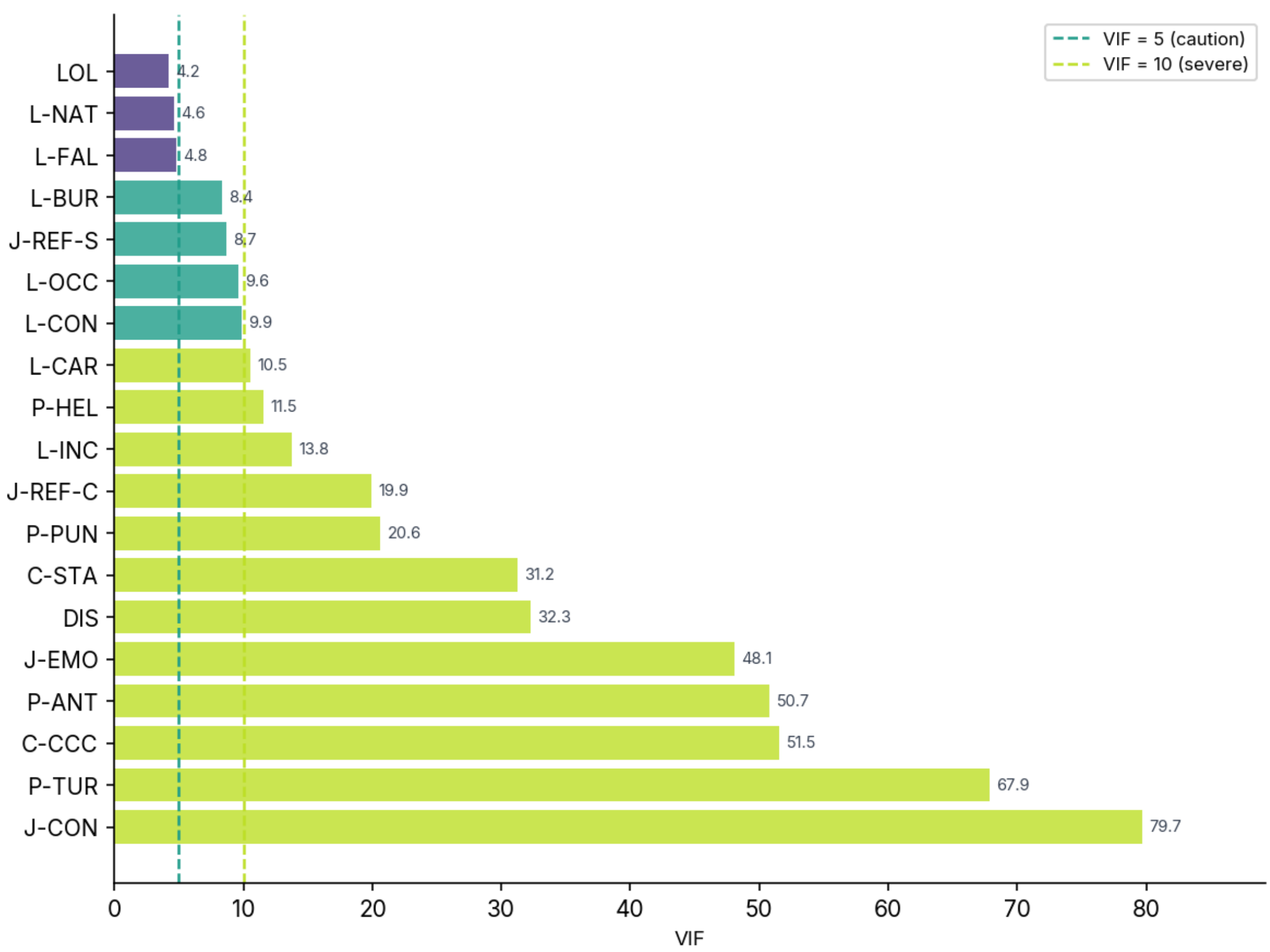

*Figure 3b. Variance inflation factors (VIF), long texts, full feature set. Orange dashed line: VIF = 5 (caution threshold); red dashed line: VIF = 10 (severe multicollinearity threshold).*

# 2. Model training

## 2.1. Feature weights and stability

Feature weight stability, assessed by re-fitting the model on each of five cross-validation folds, is displayed in Figures 4a–b. In the short text domain, the three most influential features by absolute coefficient magnitude were J-REF-S (mean = −1.093), C-CCC (+1.025), and L-INC (+0.835). The strongly negative coefficient for J-REF-S indicates that texts containing accurate information are distinguished by the presence of referential support: texts that cite and substantiate claims are substantially less likely to be classified as misinformation. C-CCC and L-INC carry the largest positive weights, confirming that causation–correlation conflation and internal inconsistency are the strongest positive fingerprints of misinformation in short texts. All coefficients showed low cross-fold standard deviation, indicating stable feature contributions.

In the long text domain, J-REF-C (−1.455) and J-EMO (+1.177) emerged as the dominant features, followed by DIS (+1.082) and P-TUR (+0.741). The strong negative weight of J-REF-C, i.e., the use of references to contest rather than support claims, in both domains (and in particular in the long text domain), may appear counterintuitive at first glance, given that higher J-REF-C scores are more common in false texts overall (Figure 1). Although negative references occur more frequently in false texts, once other hallmarks of misinformation are taken into account, the remaining signal associated with J-REF-C likely captures evidence-based critical engagement, which is characteristic of accurate long-form journalism and commentary. Further, J-EMO, the strongest positive predictor in the long text domain, further supports the observation that extended misinformation relies heavily on sustained emotional loading.

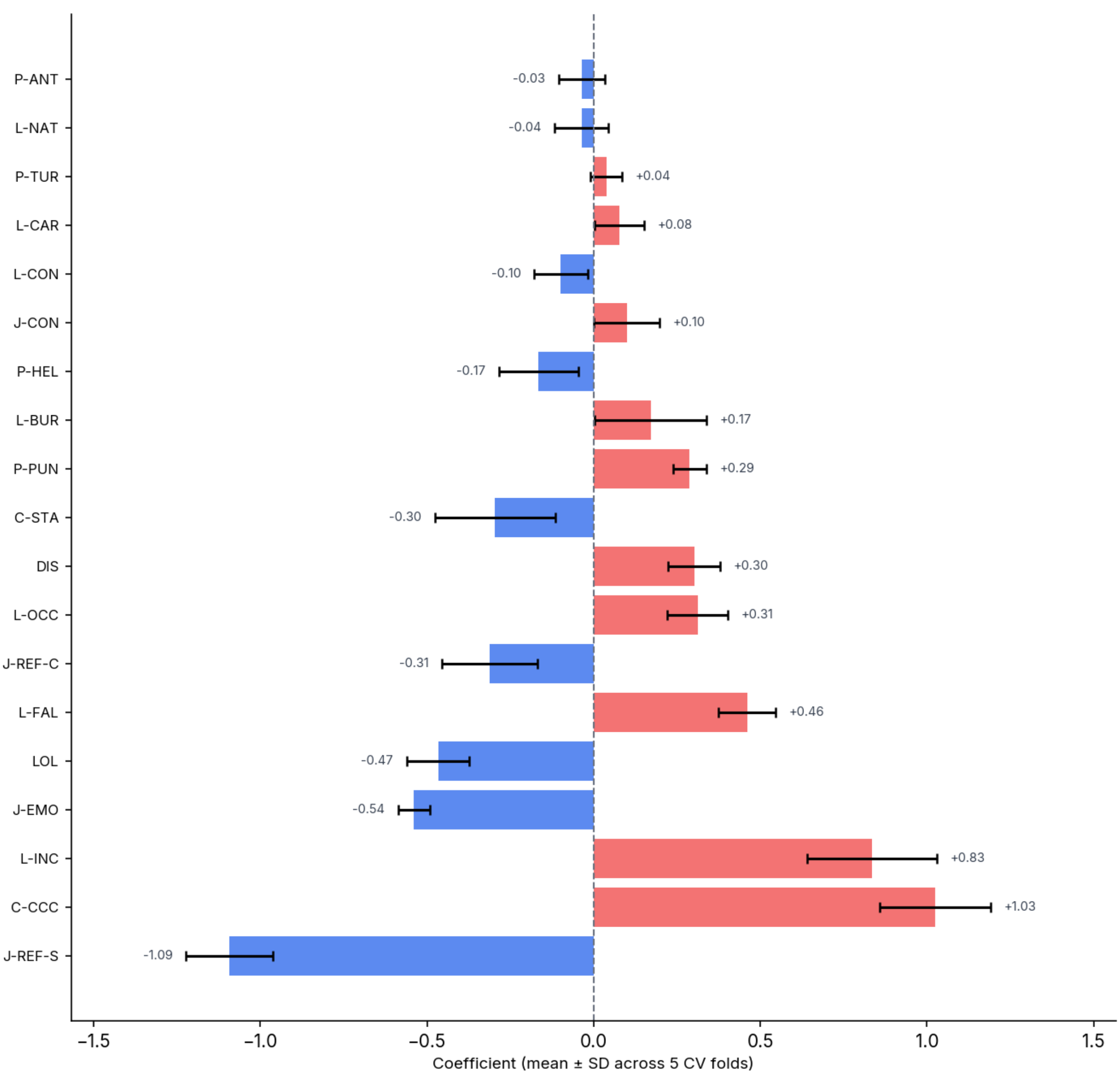

*Figure 4a. Feature weight stability, short texts. Bars show mean logistic regression coefficients across 5 cross-validation folds; error bars show ± 1 SD. Red: positive coefficients (associated with misinformation); blue: negative (associated with authentic content).*

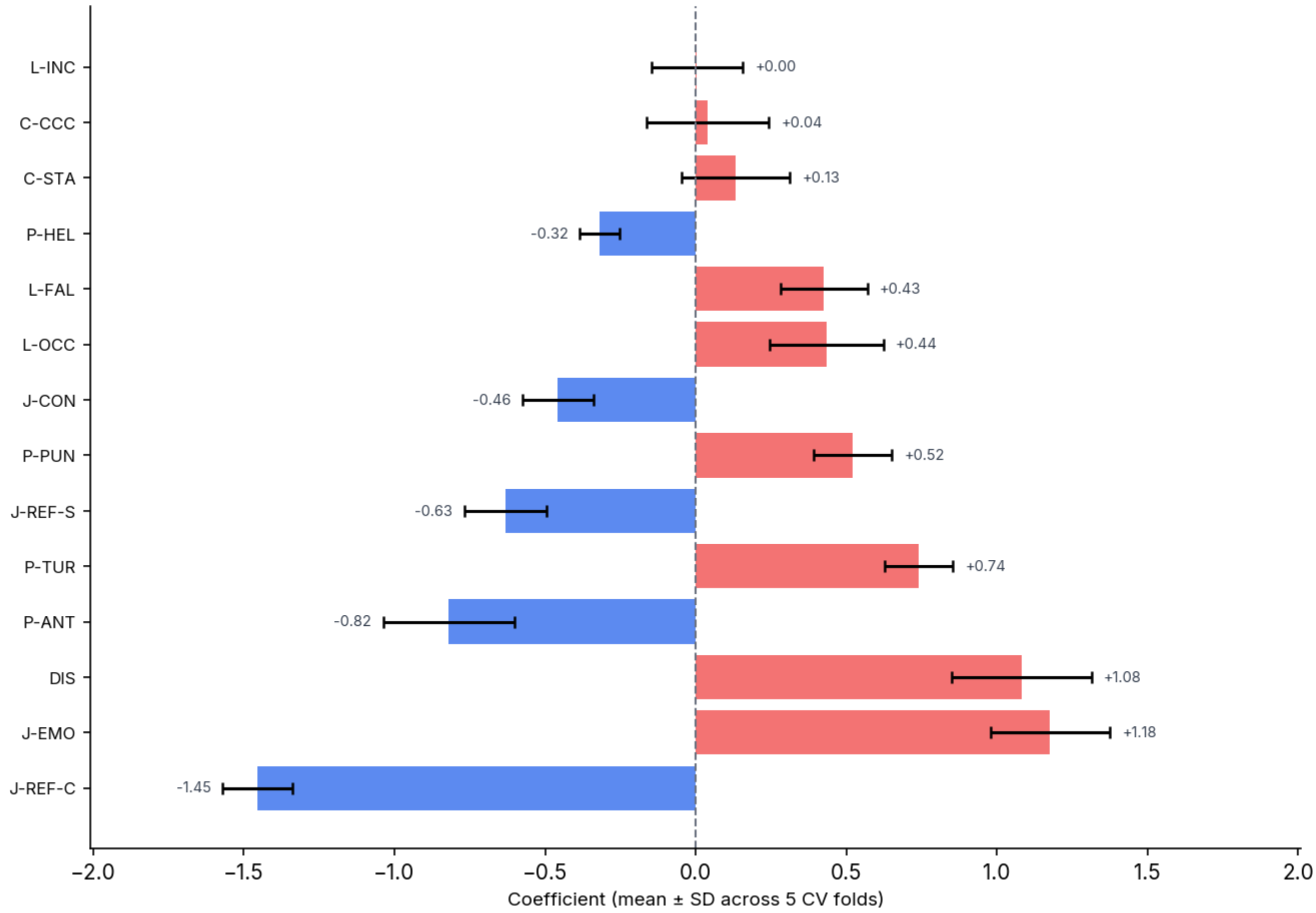

*Figure 4b. Feature weight stability, long texts. Bars show mean logistic regression coefficients across 5 cross-validation folds; error bars show ± 1 SD. Red: positive coefficients (associated with misinformation); blue: negative (associated with authentic content).*

## 2.2. Learning curves

Learning curves for both domain models are shown in Figure 5. For the short text domain, training AUC and cross-validation AUC converged rapidly, with the gap narrowing to approximately 0.02 by the full training set size. Cross-validation AUC stabilized around 0.92, indicating that the model generalizes well and that the training set size is adequate for the feature dimensionality. For the long text domain, a persistent gap between training AUC (~0.88) and cross-validation AUC (~0.84) was observed across all training sizes, consistent with greater heterogeneity in the long-text corpus. Cross-validation AUC continued to improve gradually up to the full training size, suggesting that additional long-form training examples would likely yield additional, albeit minimal, gains.

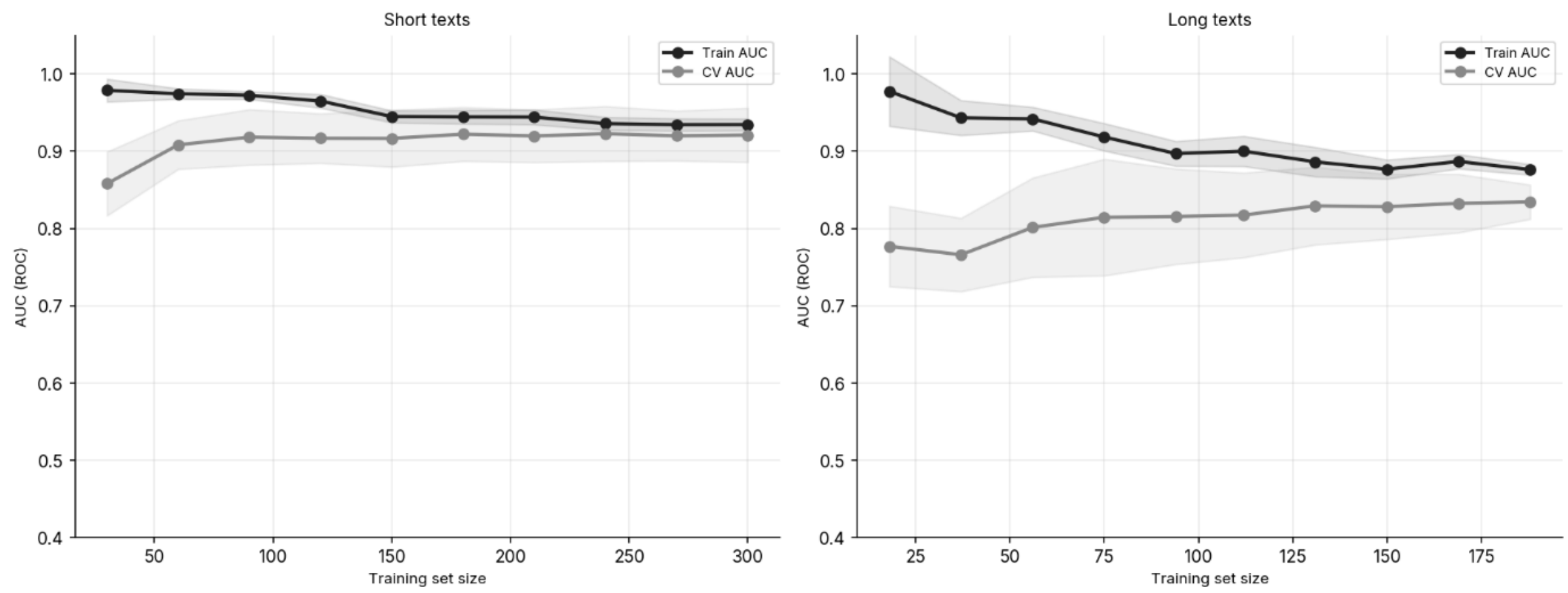

*Figure 5. Learning curves, short (left) and long (right) domain models. Train AUC (blue) and 5-fold cross-validation AUC (red) are shown as a function of training set size. Shaded bands: ± 1 SD across folds.*

# 3. Evaluation on the held-out test set

## 3.1. Discrimination and classification performance

Model performance was evaluated on the held-out test set (20% of the corpus; $N_{short\ text}$ = 95, $N_{long\ text}$ = 59). ROC and Precision-Recall curves are shown in Figure 6. The short text domain model achieved AUC-ROC = 0.863 (95% bootstrap CI: 0.780–0.934) and AUC-PR = 0.855 (baseline: 0.55). The long text domain model achieved AUC-ROC = 0.830 (95% CI: 0.703–0.933) and AUC-PR = 0.800 (baseline: 0.47). Both models substantially outperformed chance and their respective no-skill baselines across the full range of classification thresholds.

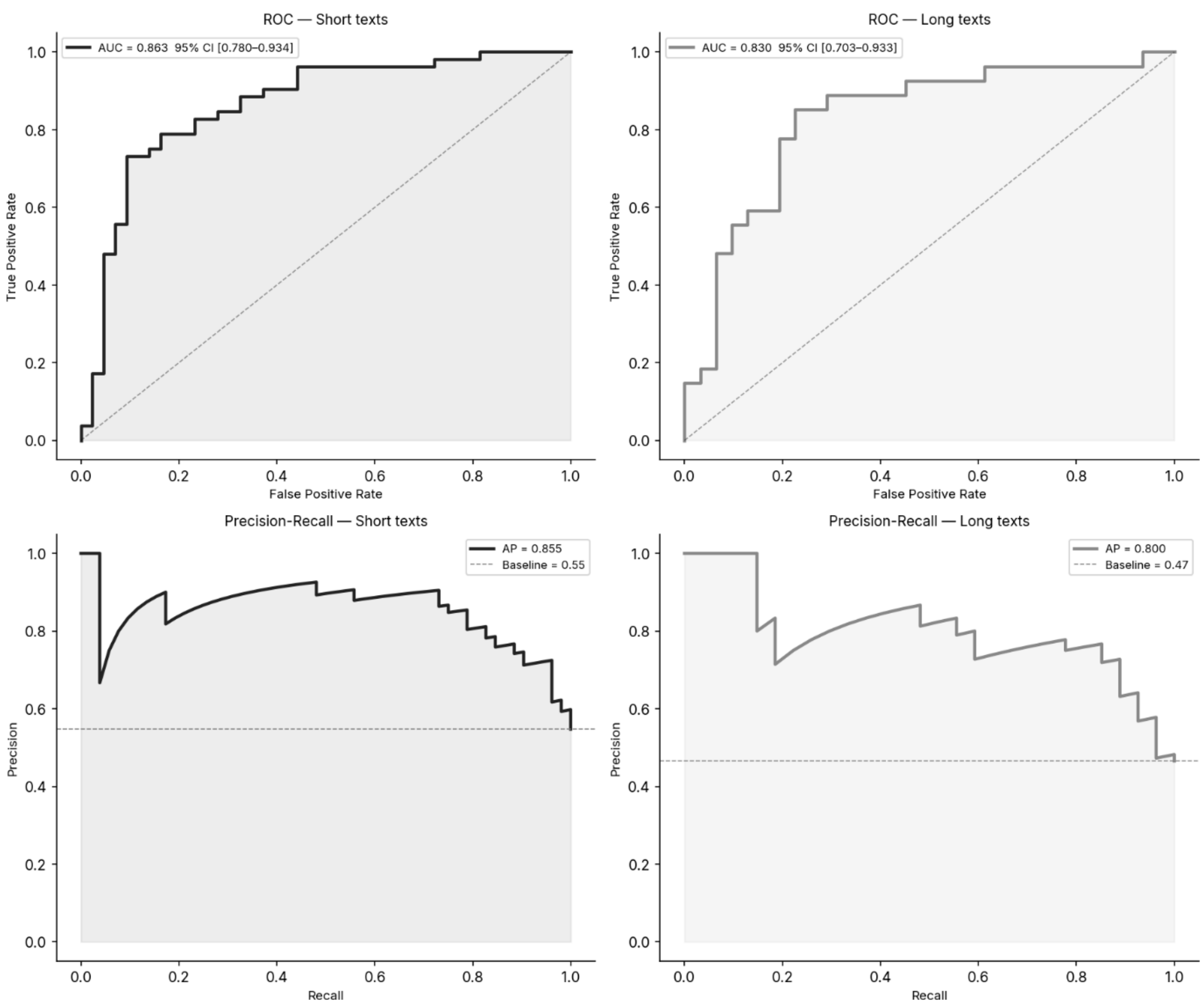

*Figure 6. ROC and Precision-Recall curves on the held-out test set. Top row: ROC curves with AUC and 95% bootstrap CI. Bottom row: Precision-Recall curves with average precision (AP) and no-skill baseline. Left column: short texts; right column: long texts.*

At the 0.50 probability threshold, classification accuracy was 78.9% (F1 = 0.788) for short texts and 79.3% (F1 = 0.793) for long texts (Table 4; Table 5). False negative rates (false information classified as true) were 21% (short texts) and 19% (long texts), while false positive rates were 21% (short texts) and 23% (long texts), indicating a symmetric error profile with no substantial asymmetry between the two error types.

| Domain | Actual | Predicted True | Predicted False | Row total |
|---|---|---|---|---|
| *Short texts* | True | 34 (79%) | 9 (21%) | 43 |
| | False | 11 (21%) | 41 (79%) | 52 |
| *Long texts* | True | 24 (77%) | 7 (23%) | 31 |
| | False | 5 (19%) | 22 (81%) | 27 |

*Table 4. Confusion matrices on the held-out test set at threshold = 0.50. Numbers indicate absolute counts; percentages are row-normalized (recall per class).*

| Metric | Short texts | Long texts |
|---|---|---|
| AUC-ROC (95% CI) | 0.863 [0.780–0.934] | 0.830 [0.703–0.933] |
| AUC-PR (average precision) | 0.855 | 0.800 |
| Accuracy | 0.789 | 0.793 |
| F1 macro | 0.788 | 0.793 |
| F1 (false class) | 0.804 | 0.786 |
| F1 (true class) | 0.773 | 0.800 |
| Brier score | 0.147 | 0.163 |
| Cohen's κ | 0.577 | 0.586 |

*Table 5. Comprehensive performance metrics on the held-out test set.*

F1 macro is the unweighted mean of per-class F1 scores and is reported as the primary classification metric because it is insensitive to class imbalance; values of 0.788 (short text domain) and 0.793 (long text domain) indicate that performance is balanced across the two classes rather than inflated by majority-class dominance. F1 scores disaggregated by class reveal a mild asymmetry: in the short text domain, the false class is classified more precisely (F1 = 0.804 vs. 0.773 for true), while in the long text domain the pattern reverses (F1 false = 0.786, F1 true = 0.800), likely reflecting the greater heterogeneity of long authentic texts drawn from both social media and PolitiFact. The Brier score, i.e., the mean squared error of predicted probabilities, quantifies calibration quality independently of threshold choice; values of 0.147 (short text domain) and 0.163 (long text domain) are well below the no-skill baseline of 0.25, indicating that the model's predicted probabilities meaningfully reflect the actual probability that a text belongs to either class rather than simply separating texts into true and false categories. Cohen's κ corrects accuracy for agreement expected by chance; κ = 0.577 (short text domain) and κ = 0.586 (long text domain) correspond to moderate-to-substantial agreement and are consistent with AUC values in the 0.83–0.86 range, indicating that the classification gains cannot be attributed to base-rate effects.

## 3.2 Score distributions and probability calibration

Score distributions and calibration plots for the test set are shown in Figure 8. Both models produced bimodal score distributions, with true texts concentrating near P(false) = 0 and misinformation texts concentrating near P(false) = 1, with greater separation in the short text domain. Reliability diagrams show that both models track the diagonal reasonably well across most of the probability range, with the short text domain model (Brier = 0.147) exhibiting slightly better calibration than the long text domain model (Brier = 0.163).

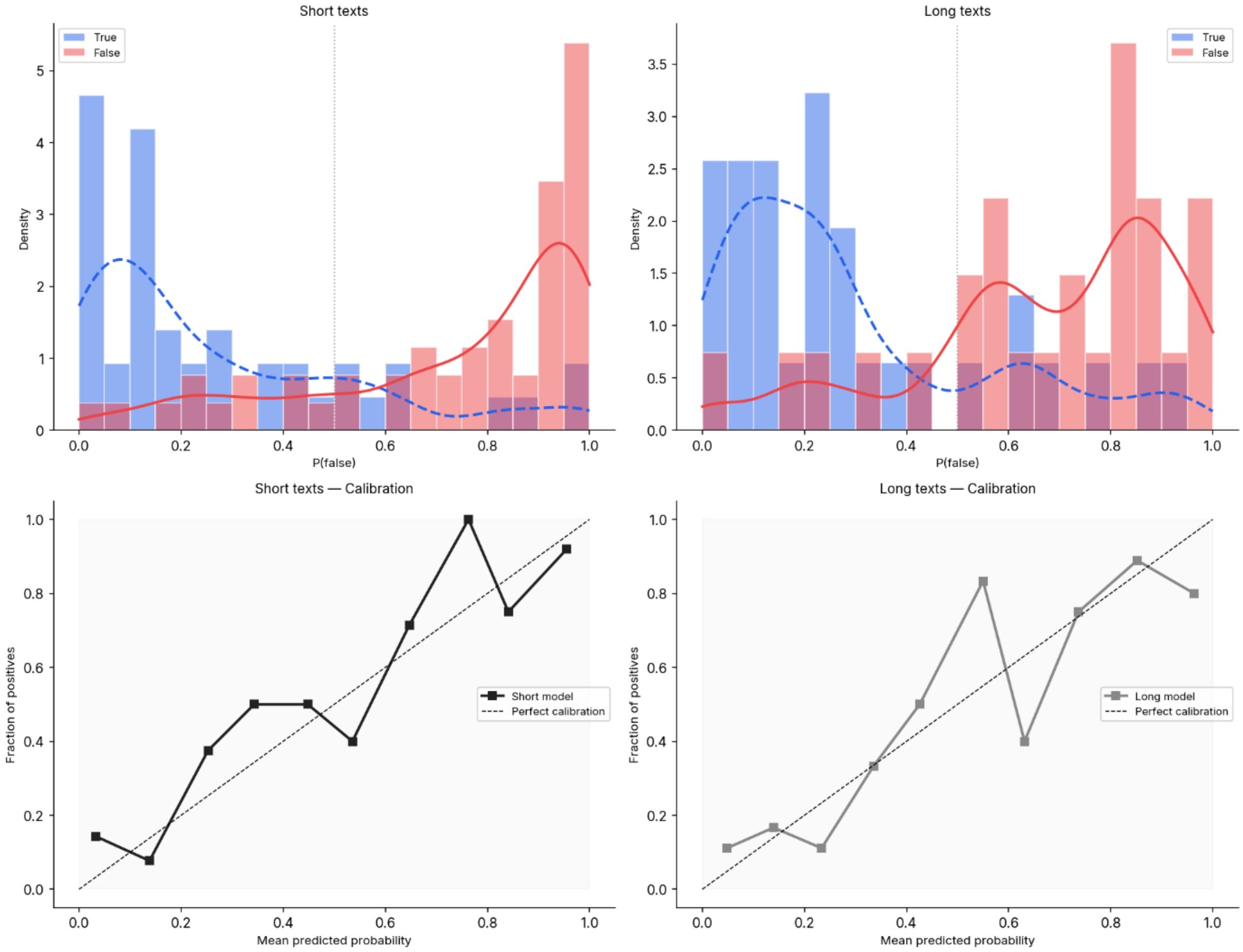

*Figure 8. Score distributions and calibration plots on the held-out test set. Top row: distribution of P(false) for true (blue) and false (red) texts; dashed vertical line at P = 0.50. Bottom row: reliability diagrams (predicted probability vs. observed fraction of positives); dashed line indicates perfect calibration. Brier scores are shown in subplot titles.*

# 4. Validation of the interpretive system

Beyond the LR classifier itself, we conducted four tests to validate the interpretive layer (risk bands, signal agreement quadrants, and caution index) as independently meaningful components of FakeSpotter's user-facing, human readable output.

## 4.1. Calibration per risk band and monotonicity

For each of the six risk bands, we computed the observed false information rate and 95% Wilson confidence interval on the scored test set (n = 153; Figure 10, test 1). Observed false information rates were: Band 0 (Very likely true information with no meaningful viral misinformation risk, n = 35): 0.09; Band 1 (Likely true information with low viral misinformation risk, n = 28): 0.32; Band 2 (Uncertain content, leaning true, with limited misinformation risk, n = 11): 0.36; Band 3 (Uncertain content, leaning misinformation, with moderate viral risk, n = 18): 0.56; Band 4 (Likely misinformation with elevated viral risk, n = 13): 0.85; Band 5 (Very likely misinformation with high viral spread potential, n = 48): 0.88.

Kendall's τ between risk band and binary label (test 2) was positive and statistically significant ($\tau = 0.559$, $p < 0.001$), confirming that the ordering of bands is monotonically associated with the probability of a text being classified as misinformation. The observed false information rates increased monotonically from Band 0 to Band 5 and fell within or closely adjacent to the expected ranges for Bands 0, 3, 4, and 5. Bands 1 and 2 showed observed false information rates slightly above their respective lower bounds, with wide confidence intervals reflecting small sample sizes. Overall, the calibration results support the claim that the risk band labels carry interpretive validity.

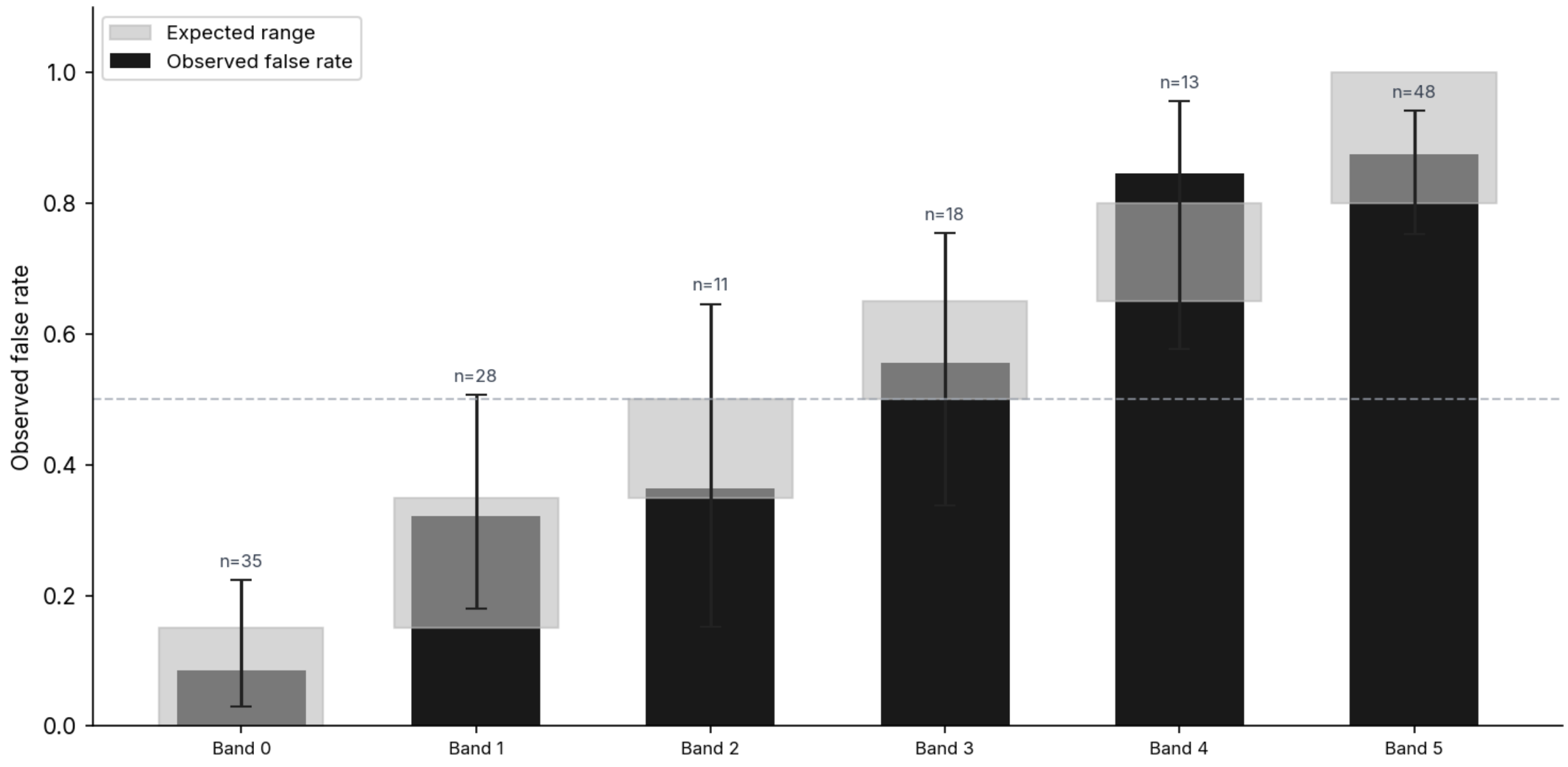

*Figure 9. Calibration per risk band. Dark gray bars: observed false information rate per band with 95% Wilson CI. Gray shading: expected false information rate range for each band. Dashed horizontal line: 50% threshold.*

## 4.2 Uncertainty honesty

Texts assigned to uncertain bands (Bands 2–3, n = 29) had substantially higher misclassification rates than texts in confident bands (Bands 0–1, 4–5, n = 124; Figure 11, test 3). A $\chi^2$ test of independence confirmed that this difference was statistically significant ($\chi^2 = 9.060$, $p = 0.0026$). This result validates the epistemic claim embedded in the band labelling: the system is calibrated to signal uncertainty precisely when it is more likely to err.

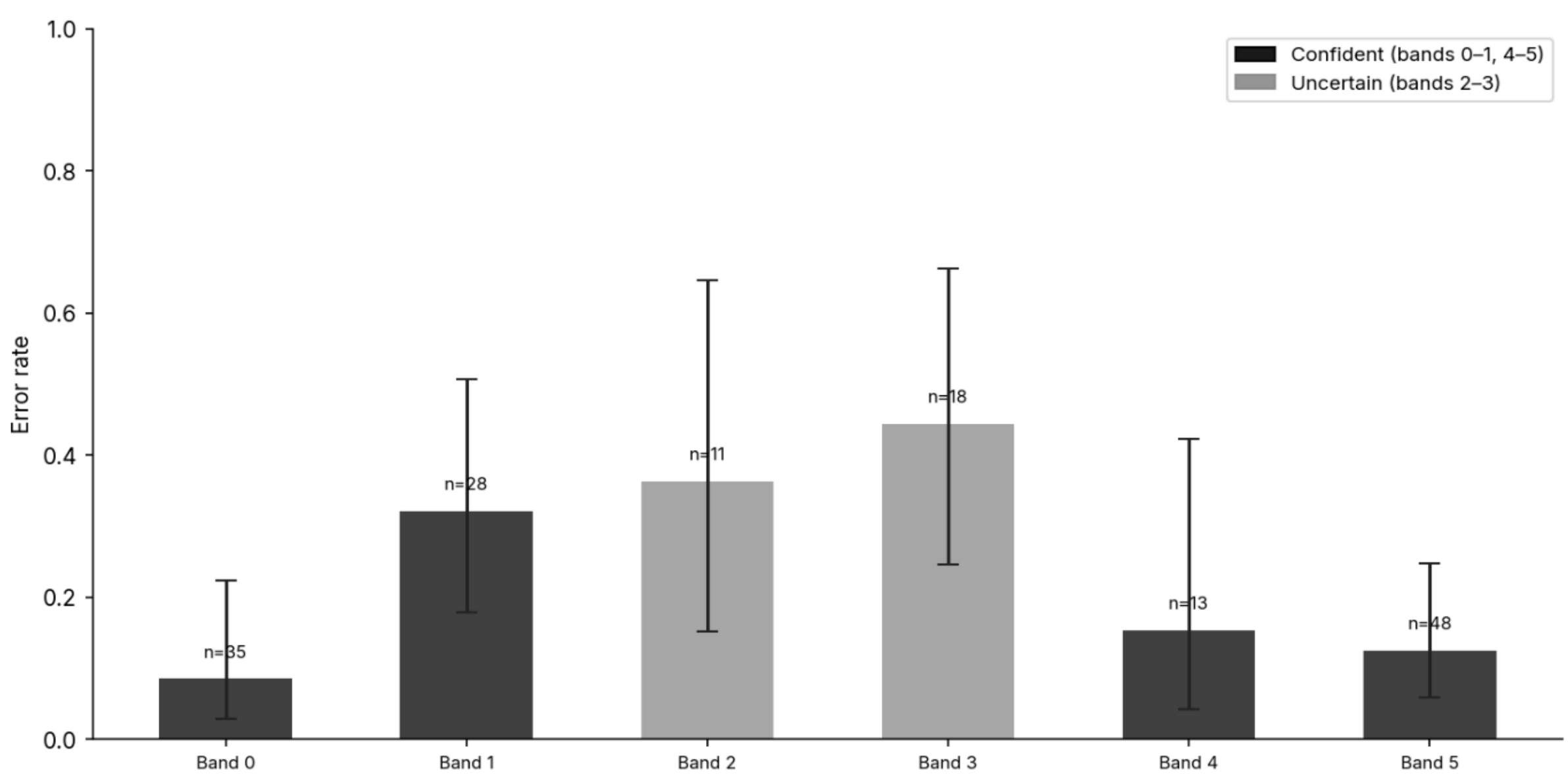

*Figure 10. Error rate per risk band. light gray bars: confident bands (0–1, 4–5); dark gray bars: uncertain bands (2–3). Error bars: 95% Wilson CI. χ² statistic and p-value for the uncertain vs. confident comparison are shown in the title.*

## 4.3. Signal agreement informativeness

Per-quadrant accuracy on the test set is shown in Figure 12 (test 4). Texts in agreement quadrants (both LR and DIS signals are either both >50 or both <50, i.e. LR and DIS-score converge towards the same interpretation of the text containing true or false information) achieved higher classification accuracy (agreement_true: 0.79, n = 64; agreement_false: 0.83, n = 58) than texts in divergence quadrants (divergence_factcheck: 0.71, n = 21; divergence_content: 0.70, n = 10). The directional pattern is consistent with the hypothesis that signal divergence carries information about interpretive uncertainty beyond what the numeric score alone conveys.

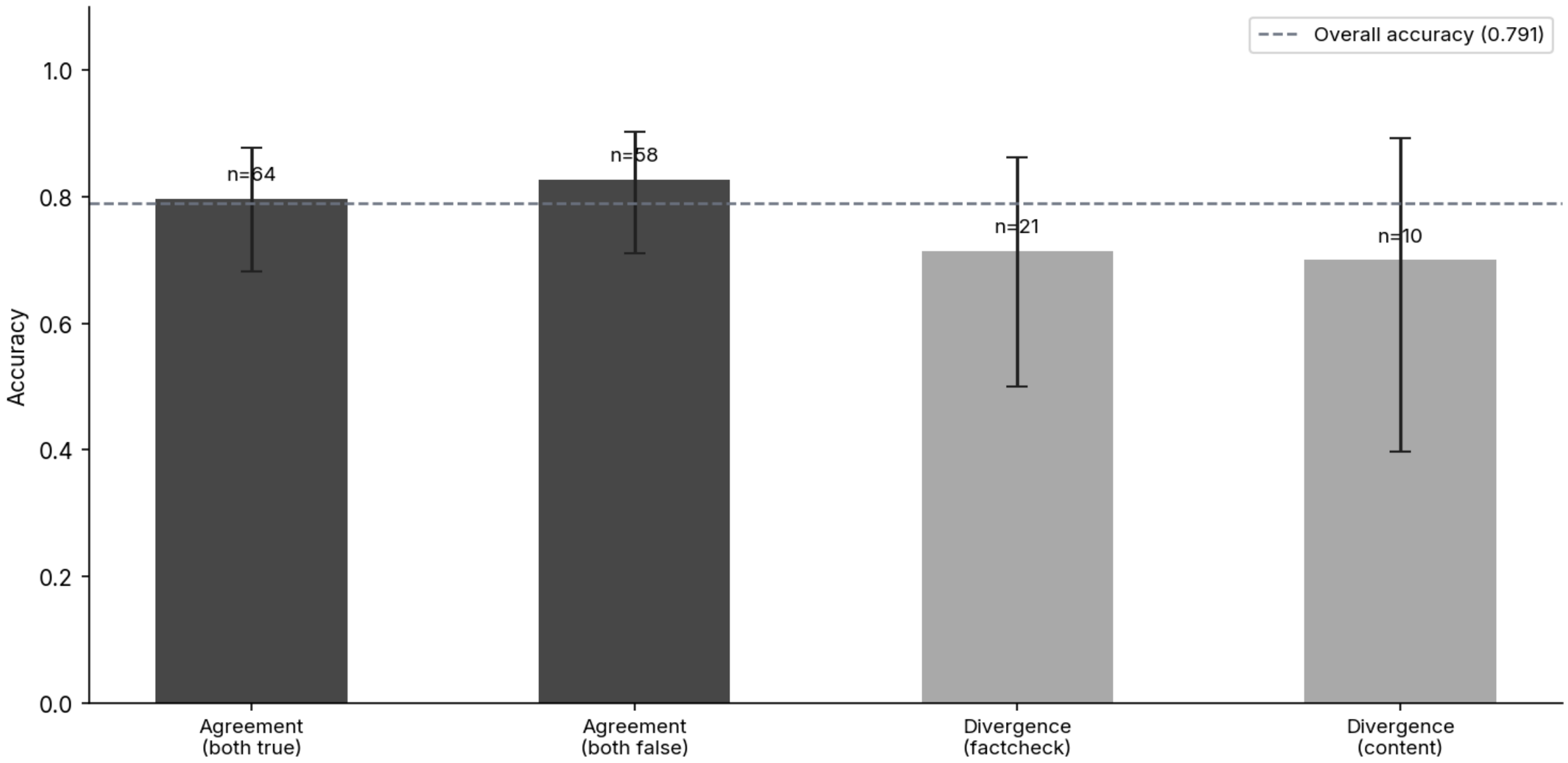

*Figure 11. Accuracy by signal_agreement quadrant. Dashed line: overall test-set accuracy (0.791). Error bars: 95% Wilson CI. χ² square test not reported due to low cell counts in divergence quadrants.*

# Discussion

## 1. Principal findings

FakeSpotter is the first misinformation detection system whose feature space is derived entirely from a pre-specified, theory-driven framework rather than from inductive feature engineering or end-to-end deep learning. The present results demonstrate that this design choice does not come at the cost of predictive performance: across both text domains, AUC-ROC values in the range 0.830-0.863 and macro F1 scores of approximately 0.79 compare favorably with published classifiers that rely on content-specific or source-based signals [32]. More importantly, they do so while operating on structural fingerprints that are, by theoretical design, content-agnostic – that is, independent of what the text is about.

Three findings are of particular conceptual significance. First, the feature weight identifies referential language – specifically, the presence of referential support (J-REF-S) and the absence of referential contestation (J-REF-C) in the analyzed texts – as the strongest discriminators of true information from misinformation with high viral potential. This is a direct empirical signature of the epistemological asymmetry between authentic communication, which must ground claims in evidence, and misinformation, which is based on persuasive narratives that are referentially impoverished. Second, causation-correlation conflation (C-CCC) and internal inconsistency (L-INC) are the dominant positive fingerprints of misinformation in short texts, while emotional loading (J-EMO) and narrative turning points (P-TUR) dominate in the long text domain. This suggests that misinformation conveyed through short texts relies on rapid inferential shortcuts and logical errors, while misinformation conveyed through longer texts sustains manipulation through narrative architecture and affective accumulation. [26]

## 2. Positioning against prior work

### 2.1. The reactive-preventive distinction

The dominant paradigm in computational misinformation detection has been reactive and content-specific: classifiers are trained to recognize known false claims, specific narrative templates, or source-level credibility signals [32,33]. This approach is effective within the distribution of training data, but it inherits the fundamental limitation identified by the theoretical framework proposed by Redaelli et al., upon which the design of FakeSpotter relies [26]: as disinformation producers adapt, new content escapes the coverage of any classifier anchored to previously observed examples. The analogy to antimicrobial resistance describes a genuine arms-race dynamic in which each content-specific detection advance invites a corresponding evasion strategy [26,34]. At the same time, new events unfold and create new demand for information, creating information voids that create the fertile ground for new misinformation to circulate [3]. This also pushes towards the creation of new false content that escapes the coverage of any classifier trained on older data.

FakeSpotter successfully operationalizes a different bet, that the structural constraints underlying persuasive manipulation—those that make a piece of information prone to virality—are more stable than the content itself [26]. The empirical results support this bet within the limits of the current corpus. Meaning: this is not a claim that the 19 dimensions are invariant across all possible forms of misinformation – it is a claim that they are stable enough to provide a useful and calibrated risk signal across the topically heterogeneous sample evaluated here, spanning public health, climate change, and political discourse across both short texts (e.g. social media posts) and long texts (e.g. media articles).

## 2.2. Comparison with analogous risk-stratification tools

A closely parallel approach to FakeSpotter's architecture has recently been proposed in the domain of nutritional misinformation. Ruani et al. developed Diet-MisRAT, a structured risk assessment tool that evaluates medium-to-long form diet and nutrition content across four risk dimensions (inaccuracy, incompleteness, deceptiveness, and health harm), yielding five-tier risk estimates rather than binary verdicts [25]. Diet-MisRAT shares with FakeSpotter several key design commitments: graded rather than binary output, explicit risk stratification validated against empirical benchmarks, and the use of a zero-shot LLM as a scoring engine under expert-designed prompts. The validation reported by Ruani et al. demonstrated strong to very strong alignment between ChatGPT-based risk estimates and expert-derived benchmarks - a finding consistent with FakeSpotter's architecture, where the LLM scores theoretically specified dimensions rather than producing unguided classifications. The domain-calibration approach of Diet-MisRAT (designed specifically for nutritional content) and FakeSpotter's content-agnostic structural fingerprinting represent complementary strategies in the broader landscape of probabilistic misinformation risk assessment.

## 2.3. LLM-mediated feature extraction

A central methodological choice in FakeSpotter is the use of an LLM (GPT-4o mini) not as a classifier, but as a feature extractor: the model is prompted to score each of the 19 theoretically specified dimensions, and these scores are then passed to a calibrated logistic regression classifier. This architecture deliberately separates semantic interpretation (the LLM's task) from statistical decision-making (the LR classifier's task). The emerging literature on LLM-based feature generation demonstrates that this approach can achieve competitive predictive performance while relying on a small set of theoretically interpretable features rather than high-dimensional embedding-based representations [35,36]. The LR classifier is fully transparent and interpretable: its coefficients are fixed, inspectable, and produce exact Shapley values. The theoretical framework is encoded in natural language, which means the scoring rubrics can be inspected and evaluated. This design contrasts with end-to-end approaches in which an LLM is fine-tuned directly for binary classification.

Furthermore, FakeSpotter mitigates the risk of hallucinated outputs by anchoring the LLM's role to a pre-specified low temperature scoring task with explicit rubrics, reducing – though not eliminating – the space for uncontrolled non-deterministic output variation. A further source of LLM-level instability is source framing. Previous work showed that when LLMs are asked to evaluate texts, their judgements remain highly consistent across models in the absence of information about the source of the text, but become systematically biased when source attribution is revealed [37]. FakeSpotter's persona by design avoids this specific failure mode by instructing the model to evaluate text content alone, providing no source information.

The multi-run averaging strategy (N = 3 independent API calls per text) addresses within-session stochasticity and provides an estimate of scoring instability via per-feature standard deviations, which feed directly into the caution index. This approach considers the evidence that LLM-based detection of linguistic features of misinformation remains sensitive to prompt variation and model version [13], and that aggregating across runs stabilizes estimates in a manner analogous to ensemble methods.

## 3. The interpretive system as an epistemic contribution

A feature of FakeSpotter that distinguishes it from most misinformation classifiers is its explicit treatment of uncertainty as a first-class output. Most binary classifiers produce a score and a threshold-determined verdict; FakeSpotter produces a score, a risk band, a caution index, and a signal agreement quadrant, each encoding a different dimension of interpretive confidence. The validation tests in Section 4 underline that texts in uncertain bands are genuinely harder to classify correctly, and texts in divergence quadrants show lower accuracy than agreement-quadrant texts, providing preliminary evidence that the caution signal adds information beyond the numeric score. This matters for two reasons: first, from a public health communication perspective, a tool that produces calibrated uncertainty estimates is substantially more valuable than one that delivers overconfident verdicts. The evidence on misinformation warning labels consistently shows that false positives - labelling true information as misinformation - erode trust in legitimate information sources through the tainted truth effect [22,23,39,40], and that prominent misinformation interventions, including fact-checking warnings, can reduce belief in false information while simultaneously increasing skepticism toward accurate content [41]. A system that honestly signals its uncertainty invites human oversight rather than displacing it. Second, the caution index provides a natural interface between the automated system and human fact-checkers or platform moderators: texts with high caution scores are the ones most likely to benefit from expert review, allowing triage of a large information stream without requiring manual examination of every item [42].

The four-quadrant signal agreement schema provides us with additional valuable information. The divergence_factcheck quadrant – in which the structural LR classifier flags false information with high virality risk while the LLM's holistic DIS score does not – identifies a pattern that may correspond to novel misinformation, i.e., misinformation not yet covered in debunking material available online and therefore absent from the model’s training data. At the same time, it may capture true information that nonetheless exhibits manipulative characteristics, such as malinformation, hate speech, harassment, or any other form of emotionally loaded, illogical, or narrative-driven content that share structural similarities with misinformation. Instead, the divergence_content quadrant – in which the DIS score flags information as false while the LR classifier does not – may identify false information with low viral or manipulative potential. In these cases, the content may be factually inaccurate, yet lack the emotionally loaded, illogical, or narrative-driven features that FakeSpotter evaluates as markers of persuasive manipulation and virality risk. Both divergence types are, by design, associated with lower classification accuracy and higher caution scores, and both warrant qualitatively different interpretive responses from users.

# 4. Limitations

## 4.1. Dataset scope, size, and generalizability

The current corpus of 764 texts, while topically diverse, is skewed toward social media posts in the short domain and PolitiFact articles in the long domain. The hand-curated short texts dataset was labelled by the research team rather than by independent annotators with measured inter-rater reliability, introducing a source of potential labelling bias that cannot be quantified post hoc. FakeNewsNet's PolitiFact subset, while a widely used benchmark, reflects the specific topical and stylistic distribution of US political fact-checking. Generalizability to other languages, other platforms (e.g., WhatsApp, Telegram, TikTok captions), or non-Western political and cultural contexts has not been established.

## 4.2. LLM dependency and adversarial fragility

FakeSpotter's scoring layer depends, in its current version, on GPT-4o mini, a proprietary model whose weights, training data, update and discontinuation schedule are not under the authors' control. Changes to the model's behavior - whether through version updates, alignment modifications, or prompt sensitivity shifts - could alter the distribution of the 19 feature scores without changing the frozen LR classifiers, potentially degrading performance in ways that are not immediately detectable, and therefore requiring re-calibration of the LR model. The multi-run averaging strategy and caution index provide partial protection against stochastic variation within a session, but they do not protect against systematic distributional shifts across model versions.

A related concern is adversarial fragility. The 19 scoring dimensions and their natural-language rubrics are described in the Methods section of this paper and available in the public repository. A sophisticated disinformation producer who has read this paper could, in principle, generate content that scores low on the features positively associated with disinformation (L-INC, C-CCC, J-EMO in the long domain) while maintaining manipulative intent, which raises some concerns about the diffusion of this information [43]. There is growing evidence that adversarial attacks targeting the specific features used by detection systems can substantially degrade classifier performance [44], and that LLMs can be used to generate fake news in real time by iteratively evading existing detectors [45]. With this said, we believe this limitation is less severe and problematic for FakeSpotter. FakeSpotter does not distinguish true from false information directly; rather, it operationalizes a theoretical framework aimed at identifying the structure features that make information persuasive, manipulative, prone to virality. As a result, successful evasion would require concealing these features while substantially altering the communicative strategies that contribute to the broad dissemination of misleading information in the first place. If such features are removed, the resulting content may become less capable of achieving persuasive and viral effects. Nevertheless, adversarial adaptation remains possible, particularly if new persuasive strategies emerge that are not captured by the current misinformation fingerprints framework [26].

### 4.3. Irony, satire, and the LOL score

The LOL score captures irony and sarcasm as a source of scoring instability, but the current implementation does not resolve the classification of ironic content. Satire, parody, and rhetorical irony share surface features with misinformation – elevated emotional loading, narrative antagonists, unfalsifiable claims – but carry opposite communicative intent. The caution index appropriately flags high-LOL texts as uncertain but provides no mechanism for distinguishing irony from misinformation in cases where other features are also elevated. This is a structural limitation of any classification approach that operates on linguistic form without access to authorial intent or broader pragmatic context.

### 4.4. The 80-word threshold

The short/long domain split at 80 words was motivated by the observed bimodality of text lengths in the corpus and validated by the distinct feature weight profiles in the two domains. However, it is a hard threshold applied at a single point, and texts near the boundary (in the range 70-90 words) may be misclassified by the wrong domain model. Smooth domain-transition strategies, such as soft blending of the two models' predictions in the boundary region, were not explored in the present study and represent a straightforward avenue for improvement.

## 5. Future directions

Several extensions of the present work are immediately tractable. First, the corpus could be expanded, with particular emphasis on long-domain texts and on content from non-Anglophone contexts. The learning curves suggest that the long-domain model has not reached performance saturation; additional data in this domain would likely close the residual train-validation gap. Further, the current system scores texts at the point of submission, without access to the broader information context: the source, the publication venue, the response it has elicited, or the network through which it circulates. In addition, the theoretical framework based on which FakeSpotter is built should be stress-tested against adversarially generated content. LLMs are now capable of producing persuasive disinformation that adopts the surface conventions of credible communication [46,47], and evaluating FakeSpotter's performance on such content would allow us to establish the limits of the content-agnostic approach to misinformation detection and identify which classes of misinformation are most susceptible to evasion.

Finally, and most broadly, FakeSpotter's interpretive architecture – probabilistic risk bands, explicit caution signals, and a four-quadrant disagreement schema – opens the possibility of human-supervised workflows in which the system identifies and triages uncertain cases for expert review rather than delivering terminal verdicts. The evidence from explainability research on misinformation detection suggests that confidence scores and explicit explanations, when designed carefully, can support more calibrated human decision-making [42], though the same literature also cautions that explainability features can introduce uncertainty and reduce classification agreement if not integrated thoughtfully.

# Conclusion

The central wager of FakeSpotter is that the structural grammar of misinformation is more stable than its lexicon. The results reported here provide initial empirical support for this claim: a system trained on 19 theoretically motivated dimensions of linguistic, narrative, logical, and critical-reasoning structure achieves AUC-ROC values of 0.830-0.863 on a topically heterogeneous test corpus, with calibrated probability outputs and an interpretive layer that honestly communicates its own uncertainty. These properties – theoretical grounding, interpretability, calibration, and epistemic honesty – do not, individually or collectively, solve the misinformation problem. But they constitute a different kind of contribution than a classifier optimized for a single benchmark: they provide a reusable, extensible, and theoretically accountable scaffold for the long-term task of building information ecosystems that are resilient to manipulation, rather than merely reactive to current forms of misinformation [26].

# Ethical considerations

The hand-curated dataset was assembled from publicly available social media posts. To prevent back-tracing, re-identification, or potential stigma, all original links were retained internally but are not published; examples presented in associated publications have been anonymized and lightly paraphrased. No personal data was collected or stored.

FakeSpotter is a probabilistic tool and does not produce determinate verdicts. The risk index and risk bands are calibrated estimates, not classifications. The caution index and signal agreement quadrant are designed to make the limits of each assessment transparent to the end user. We explicitly discourage uses of FakeSpotter that would reduce its probabilistic output to binary, consequential decisions without human oversight.

## 1. Code availability and responsible release

In accordance with our broader view that openness in science should be understood as an instrumental value, to be balanced against other ethically relevant considerations [43], we do not adopt a policy of unrestricted full release for all components of FakeSpotter at this stage. The system presents a foreseeable dual-use risk: because it is designed to detect disinformation through recurrent structural fingerprints rather than topic-specific content, full disclosure of its most operationally sensitive components could potentially facilitate adversarial adaptation. In particular, releasing fully reusable prompts, detailed scoring rubrics, calibrated model artifacts, decision thresholds, feature-weighting specifications, and other inference-enabling implementation details could enable malicious users to reverse-engineer the system's detection logic and iteratively optimize manipulative content to evade it.

## 2. Access to restricted materials

Restricted components may be made available under controlled conditions for legitimate research purposes, subject to case-by-case review, a stated research objective, and agreement not to redistribute or use the material to develop evasion strategies or manipulative communication tools.

# Supplementary materials

## Supplementary results

### Univariate distributions of the FakeSpotter indicators by veracity class

These univariate distributions provide the feature-level basis for the multivariate, per-domain models reported in the main text. In the short-text domain all 19 indicators separate the two classes significantly after Bonferroni correction (18 at p < 0.001; J-EMO at p < 0.01), with several large effects: the strongest separators are L-INC (r_rb = 0.71, d = 1.50), C-CCC (0.67, 1.39) and DIS (0.62, 1.29), all elevated in false texts, while J-REF-S (r_rb = −0.47, d = −1.09) and P-HEL (−0.42, −0.86) are instead elevated in true texts. In the long text domain the same indicators dominate but effect sizes are markedly attenuated (max |d| ≈ 0.98): DIS (0.51, 0.98), J-REF-S (−0.47, −0.88) and C-CCC (0.46, 0.92) lead, and five indicators (L-BUR, LOL, L-CON, L-NAT, L-CAR) no longer reach significance. Two regularities hold across domains: separation is systematically stronger for short texts (all 19 significant, max rrb = 0.71) than for long texts (14 significant, max rrb = 0.51), and the direction of every significant difference is consistent — J-REF-S and P-HEL track truthful content, the remaining indicators track false content. This pattern justifies modelling the two domains separately and motivates the feature-selection step, since the long text domain carries redundant and uninformative indicators that the reduced models can safely discard. As univariate comparisons, these tests ignore inter-feature correlation (see VIF, Fig 3a and 3b) and are reported for transparency rather than as the selection criterion.

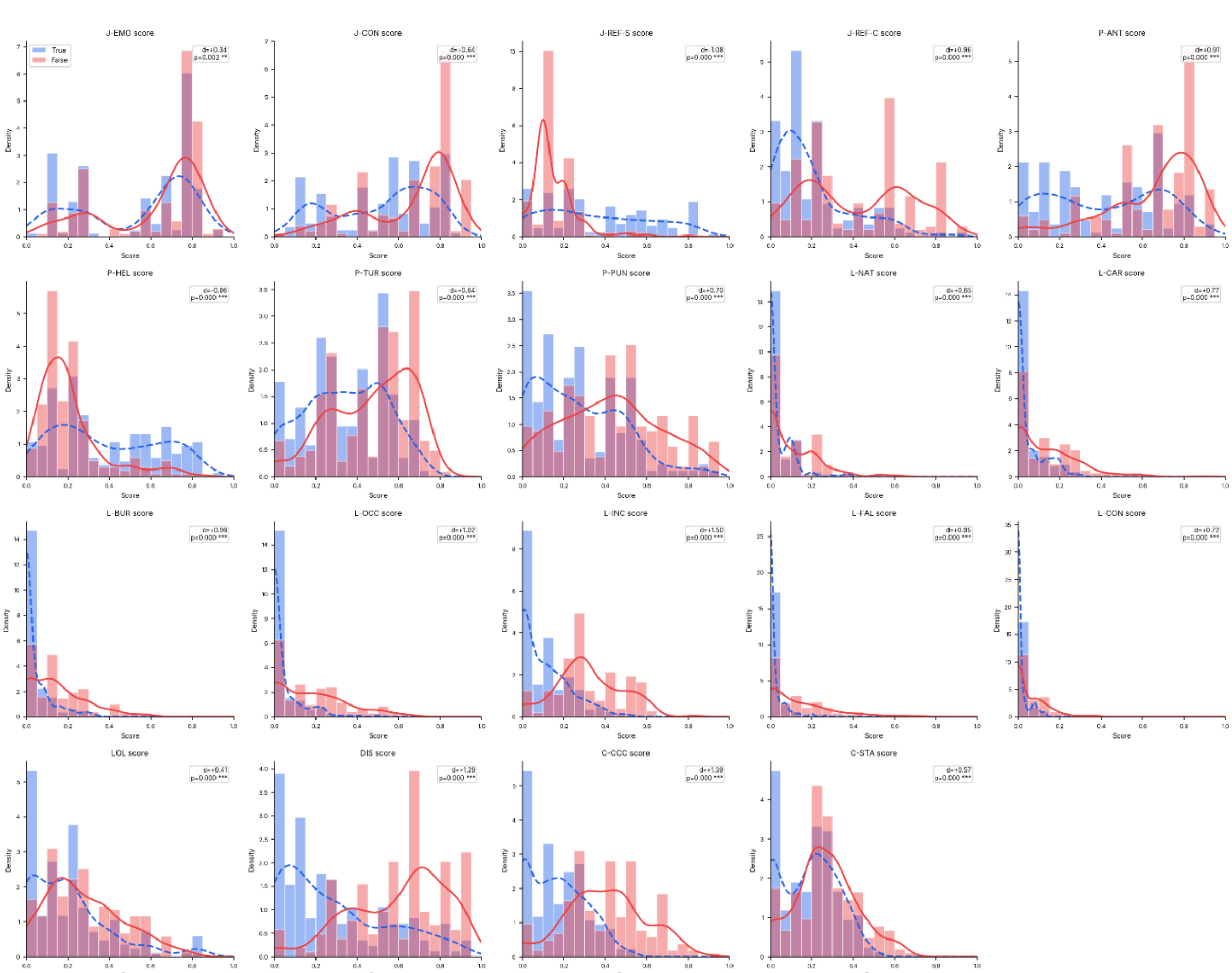

*Figure S1a. Univariate score distributions of the 19 FakeSpotter indicators by veracity class — short texts (≤ 80 words, training set). For each indicator, mean assessment scores (range 0–1) are shown as normalized histograms (21 bins) overlaid with Gaussian kernel-density estimates, separately for items labelled false (red) and true (blue). Each panel reports the standardised mean difference (Cohen's d; positive = higher in false items)*

*and the two-sided Mann–Whitney U p-value after Bonferroni correction across the 19 indicators (*** $p < 0.001$, ** $p < 0.01$, * $p < 0.05$, ns = not significant); indicators are ordered by absolute rank-biserial correlation.*

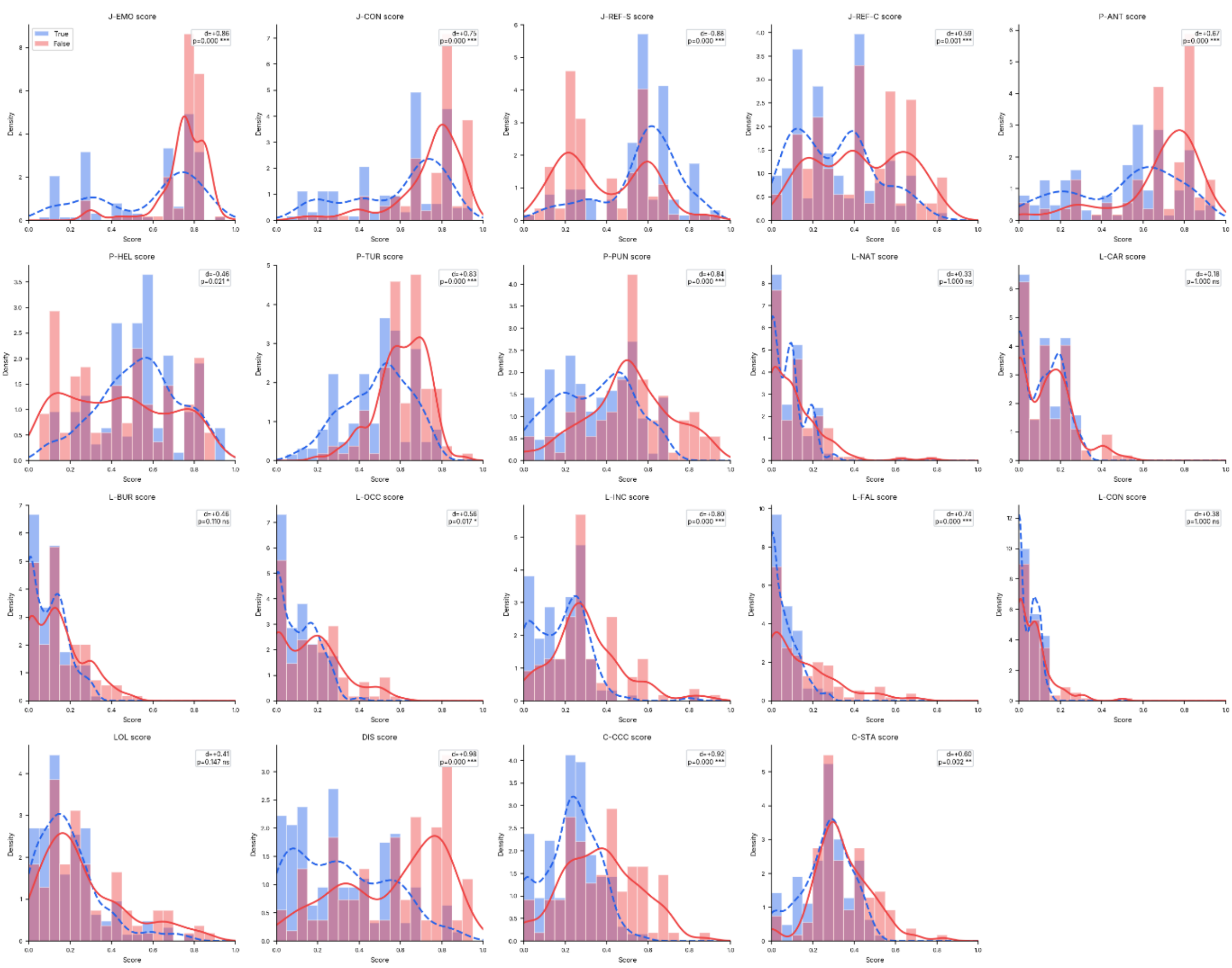

*Figure S1b. Univariate score distributions of the 19 FakeSpotter indicators by veracity class — long texts.*

## Feature importance: SHAP analysis

SHAP (SHapley Additive exPlanations) values quantify the contribution of each feature to an individual prediction by distributing the difference between the model output and its expected value across all input features, in a manner consistent with game-theoretic fairness axioms. For logistic regression, LinearExplainer computes exact Shapley values without approximation. A positive SHAP value indicates that a feature pushed the prediction toward misinformation; a negative value indicates a push toward true information. Features are ranked by mean |SHAP| across the test set, providing a global importance measure that integrates both the magnitude and the frequency of each feature's contributions.

SHAP values were computed for all test-set texts in each domain (Figures 9a–b). In the short text domain, the top contributors by mean |SHAP| were C-CCC, J-REF-S, L-INC, J-EMO, and LOL. High C-CCC scores drove predictions strongly toward false information (SHAP > +2.5 in some cases), while high J-REF-S scores — reflecting the presence of referential support — exerted the largest negative SHAP values, pulling predictions toward true information.

In the long text domain, J-REF-C dominated by mean |SHAP|, followed by J-EMO, DIS, P-ANT, and P-TUR. High J-REF-C scores (heavy contestation of references) pushed

predictions strongly toward true information — again coherent with the weight-stability results — while high J-EMO and DIS scores pushed strongly toward misinformation. Notably, the SHAP distributions for P-TUR in the long domain showed that a high turning-point score was almost universally associated with a positive contribution to the misinformation prediction, consistent with the theoretically expected role of narrative escalation in extended misinformation.

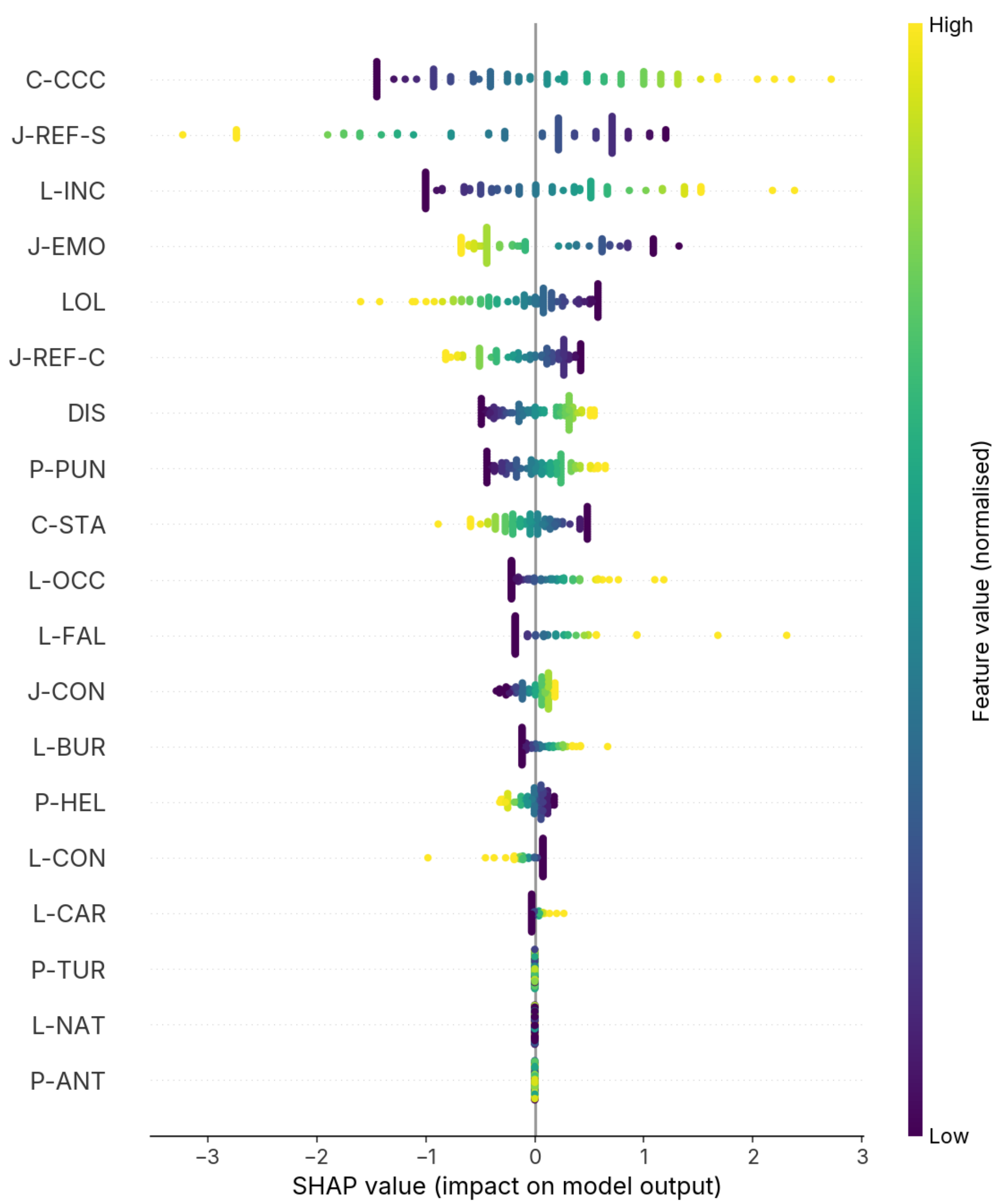

*Figure S2a. SHAP summary plot — short texts, test set. Each point represents one test text. X-axis: SHAP value (contribution to log-odds of P(false)); color: normalized feature value (red = high, blue = low). Features sorted by mean |SHAP|.*

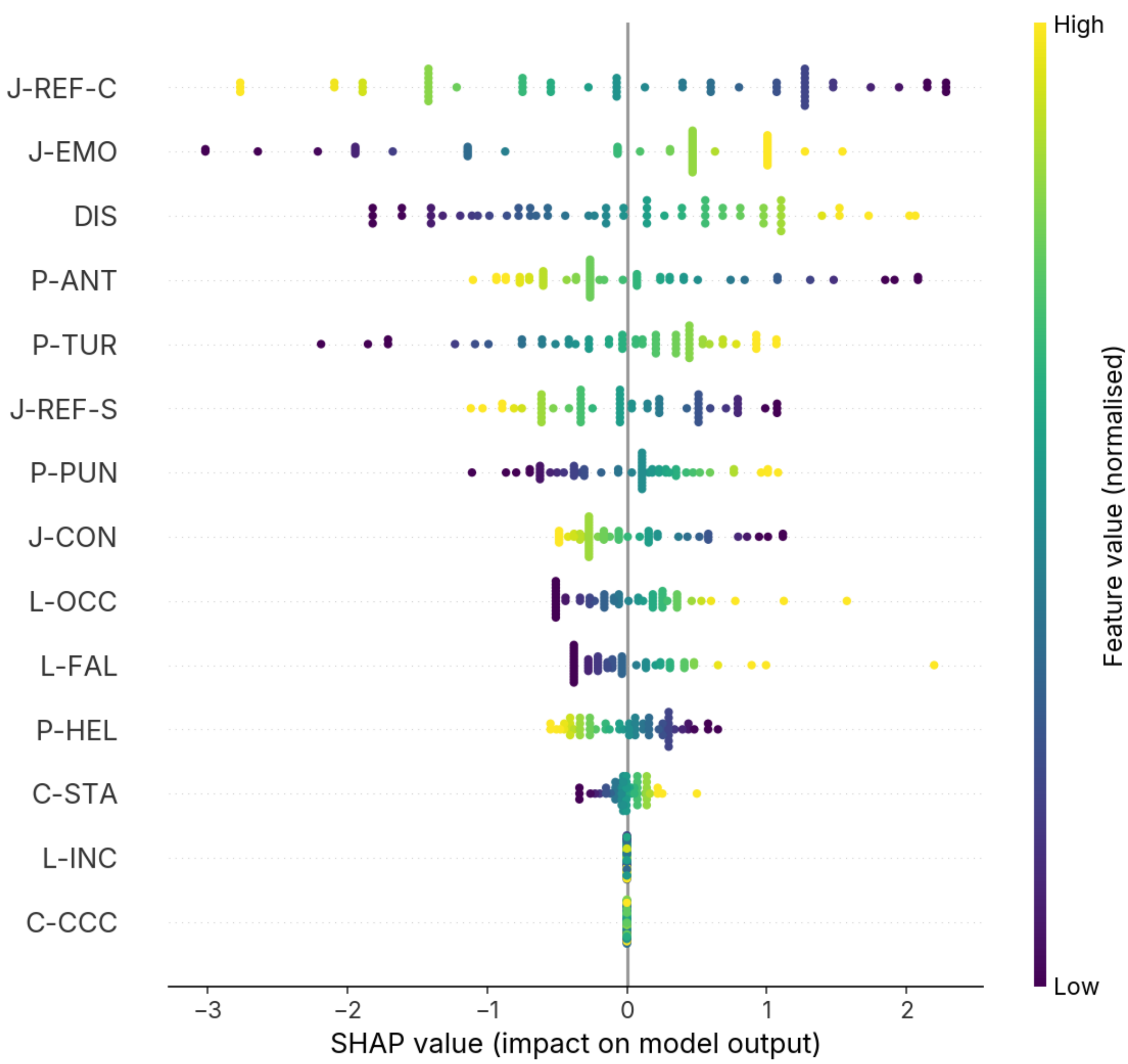

*Figure S2b. SHAP summary plot — long texts, test set.*